\documentclass{article}

\usepackage[final]{neurips_2025}

\usepackage[utf8]{inputenc} % allow utf-8 input
\usepackage[T1]{fontenc}    % use 8-bit T1 fonts
\usepackage{hyperref}       % hyperlinks
\usepackage{url}            % simple URL typesetting
\usepackage{booktabs}       % professional-quality tables
\usepackage{amsfonts}       % blackboard math symbols
\usepackage{nicefrac}       % compact symbols for 1/2, etc.
\usepackage{microtype}      % microtypography
\usepackage{xcolor}         % colors

\usepackage{amsmath}
\usepackage{amssymb}
\usepackage{newfloat}
\usepackage{verbatim}
\usepackage{multirow}
\usepackage{boldline}
\usepackage{subcaption}
\usepackage{hyperref}
\usepackage{marvosym}
\usepackage{threeparttable}
\usepackage{lipsum}
\usepackage{adjustbox}
\usepackage{colortbl}
\usepackage[hang,flushmargin]{footmisc}

\usepackage{tikz}

\DeclareFloatingEnvironment[name={Supplementary Figure}]{suppfigure}
\DeclareFloatingEnvironment[name={Supplementary Table}]{supptable}

\newcommand{\xmli}[1]{{\color[rgb]{1,0.0,0.1}{[XM:#1]}}}

\newcommand{\methodname}{MePH}

\title{AI-Enabled Accurate Non-Invasive Assessment of Pulmonary Hypertension Progression via Multi-Modal Echocardiography}

\vspace{-10pt}
\author{%
  Jiewen Yang$^1$, Taoran Huang$^2$, Shangwei Ding$^3$, Xiaowei Xu$^{2,6}$, Qinhua Zhao$^4$, \\ 
  \AND
  Yong Jiang$^{5,7}$, Jiarong Guo$^1$, Bin Pu$^1$, Jiexuan Zheng$^2$, 
  \AND
  Caojin Zhang$^2$, Hongwen Fei$^2$, Xiaomeng Li$^{1,8}$
}

\begin{document}

\footnotetext[1]{Department of Electronic and Computer Engineering, The Hong Kong University of Science and Technology, Hong Kong SAR}
\footnotetext[2]{Guangdong Cardiovascular Institute, Guangdong Provincial People’s Hospital (Guangdong Academy of Medical Sciences), Southern Medical University, Guangzhou, Guangdong Province, China}
\footnotetext[3]{Department of Ultrasound, The First Affiliated Hospital of Guangzhou Medical University, Guangzhou, Guangdong, China}
\footnotetext[4]{Department of Pulmonary Circulation, Shanghai Pulmonary Hospital, Tongji University School of Medicine, Shanghai, China}
\footnotetext[5]{Department of Echocardiography, Fuwai Hospital Chinese Academy of Medical Sciences, Shenzhen, China}
\footnotetext[6]{Guangdong Provincial Key Laboratory of South China Structural Heart Disease, Guangzhou, China}
\footnotetext[7]{State Key Laboratory of Cardiovascular Disease, Department of Echocardiography, National Center for Cardiovascular Diseases, Fuwai Hospital, Chinese Academy of Medical Sciences and Peking Union Medical College, Beijing, China}
\footnotetext[8]{Department of Computer Science and Engineering, The Hong Kong University of Science and Technology, Hong Kong SAR}

\maketitle

\vspace{-10pt}
\begin{abstract}
\vspace{-10pt}
Echocardiographers can detect pulmonary hypertension using Doppler echocardiography; however, accurately assessing its progression often proves challenging. Right heart catheterization (RHC), the gold standard for precise evaluation, is invasive and unsuitable for routine use, limiting its practicality for timely diagnosis and monitoring of pulmonary hypertension progression.
Here, we propose MePH, a multi-view, multi-modal vision-language model to accurately assess pulmonary hypertension progression using non-invasive echocardiography. 
We constructed a large dataset comprising paired standardized echocardiogram videos, spectral images and RHC data, covering 1,237 patient cases from 12 medical centers.
For the first time, MePH precisely models the correlation between non-invasive multi-view, multi-modal echocardiography and the pressure and resistance obtained via RHC.
We show that \methodname{} significantly outperforms echocardiographers' assessments using echocardiography, reducing the mean absolute error in estimating mean pulmonary arterial pressure (mPAP) and pulmonary vascular resistance (PVR) by 49.73\% and 43.81\%, respectively.
In eight independent external hospitals, \methodname{} achieved a mean absolute error of 3.147 for PVR assessment.
Furthermore, \methodname{} achieved an area under the curve of 0.921, surpassing echocardiographers (area under the curve of 0.842) in accurately predicting the severity of pulmonary hypertension, whether mild or severe. A prospective study demonstrated that MePH can predict treatment efficacy for patients.
Our work provides pulmonary hypertension patients with a non-invasive and timely method for monitoring disease progression, improving the accuracy and efficiency of pulmonary hypertension management while enabling earlier interventions and more personalized treatment decisions.
\end{abstract}

\section*{Introduction}

% Version by li 

Pulmonary Hypertension (PH) is a chronic and progressive cardiovascular condition that, if left untreated, can lead to right heart failure and ultimately result in death. It affects approximately 70 million people worldwide, accounting for nearly 1\% of the global population~\cite{hoeper2013definitions,humbert20222022,moreira2015prevalence,hoeper2016global,anderson2022pulmonary}.
Right heart catheterization (RHC) is considered the gold standard for diagnosing PH, as it measures mean Pulmonary Artery Pressure (mPAP) and Pulmonary Vascular Resistance (PVR)~\cite{hoeper2013definitions,chatterjee2009swan,nossaman2010history,humbert20222022}. Patients are typically advised to undergo re-evaluation every 18-24 months using RHC, and if there are signs of disease progression, a follow-up RHC should be conducted without delay~\cite{ishii2023prognostic,boucly2024risk}.
However, the invasive nature of these procedures and the risk of catheter-related complications raise significant concerns~\cite{wyman1988current,hoeper2006complications,al2019safety,vitiello1998complications}, making repeated assessments impractical for ongoing monitoring of cardiac function in patients with PH.

Echocardiography is recommended as the primary non-invasive modality for the initial screening of suspected PH~\cite{humbert20222022,d2022echocardiographic,noordegraaf2019pathophysiology}. It effectively assesses right ventricular morphology and function~\cite{yamasaki2016clinical,schneider2022severe,mouratoglou2020duration,humbert20222022,d2022echocardiographic,noordegraaf2019pathophysiology}, as well as estimating hemodynamic parameters~\cite{berger1985quantitative,currie1985continuous,chemla2004new,fisher2007estimating,swift2013noninvasive,abbas2013noninvasive}.
By employing Doppler imaging to calculate hemodynamic measurements and flow assessment using the modified Bernoulli equation, echocardiographers can achieve high accuracy in differentiating between PH and non-PH conditions~\cite{fisher2007estimating,swift2013noninvasive,abbas2013noninvasive}. However, the performance of these methods when assessing mPAP and PVR—both crucial for evaluating PH—has been unsatisfactory.
%Specifically, the mean absolute error (MAE) is 9.609 (mmHg) in estimating mPAP for PH diagnosis, 2.796 (WU) for PVR assessment, and overall accuracy is 37.27\% for treatment efficacy evaluation, as shown in Figure~\ref{fig:diagnosis_result_mpap}a, Figure~\ref{fig:diagnosis_result_mpap}c, and Figure~\ref{fig:diagnosis_result_pvr}c.

Deep learning methods have seen growing application in cardiovascular disease detection~\cite{wang2024screening,christensen2024vision} and cardiac function assessment~\cite{he2023blinded,attia2019screening,hannun2019cardiologist,ouyang2020video,christensen2024vision}. While some approaches aim to improve the diagnostic performance of PH using echocardiography~\cite{liao2023automatic,kwon2020artificial,diller2022framework,ragnarsdottir2024deep,sun2024chamber}, they mainly rely on a single modality, such as echocardiographic images, without integrating RHC data. This limitation reduces their effectiveness, as these methods mainly aim to match—rather than exceed—the diagnostic accuracy of expert echocardiographers. Consequently, they are insufficient for predicting hemodynamic parameters, such as mPAP and PVR. Moreover, there is no systematic, multi-center evaluation of these approaches for PH diagnosis, severity grading, or prospective studies on predicting treatment outcomes in clinical practice.

% these methods fail to model the correlation between multi-modal, multi-view echocardiographic data and the gold-standard RHC, thereby limiting their ability to accurately predict mPAP and PVR. 

% However, these methods rely solely on a single imaging modality for classifying PH and non-PH cases, without incorporating or modeling the essential hemodynamic parameters—mPAP and PVR—which are foundational to the clinical diagnosis of PH. This lack of integration limits their physiological interpretability and diagnostic robustness. 

% (1) These methods focus solely on classifying PH and non-PH cases without evaluating key hemodynamic parameters such as mPAP and PVR, which are essential for assessing PH severity.
% (2) These methods rely on either echocardiogram images~\cite{diller2022framework,sun2024chamber} or videos~\cite{liao2023automatic,ragnarsdottir2024deep}, without considering the multi-view and multi-modality aspects, which include echocardiogram videos (showing dynamic cardiac motion), Doppler images (providing hemodynamic information), and metadata (indicating gender and age). Therefore, this limitation results in an incomplete understanding of the varying levels of mPAP and PVR, leading to limited results.
% (3) Prior work has not undergone systematic multi-center evaluations for PH recognition, severity assessment, and effectiveness in predicting treatment efficacy, highlighting their limited applicability in real clinical settings.

\if 1 
While deep learning methods have been developed to improve the diagnostic performance of PH from echocardiography~\cite{liao2023automatic,kwon2020artificial,diller2022framework,ragnarsdottir2024deep,sun2024chamber}, several limitations remain:
(1) These methods focus solely on classifying PH and non-PH cases without evaluating key hemodynamic parameters such as mPAP and PVR, which are essential for assessing PH severity.
(2) These methods rely on either echocardiogram images~\cite{diller2022framework,sun2024chamber} or videos~\cite{liao2023automatic,ragnarsdottir2024deep}, without considering the multi-view and multi-modality aspects, which include echocardiogram videos (showing dynamic cardiac motion), Doppler images (providing hemodynamic information), and metadata (indicating gender and age). Therefore, this limitation results in an incomplete understanding of the varying levels of mPAP and PVR, leading to limited results.
(3) Prior work has not undergone systematic multi-center evaluations for PH recognition, severity assessment, and effectiveness in predicting treatment efficacy, highlighting their limited applicability in real clinical settings.
\fi 
 
In this paper, we construct the first large-scale dataset comprising 1,237 cases of PH patients collected from 12 medical centers. Each patient case includes multi-view, multi-modality echocardiogram videos, Doppler images, and metadata extracted from electronic health records. All patients underwent RHC to obtain measurements of mPAP and PVR.
To accurately model the correlation between echocardiography and RHC for predicting mPAP and PVR, we propose MePH—an innovative multi-view, multi-modal echocardiography vision-language model for PH assessment.
MePH accurately predicts mPAP and PVR in alignment with RHC by systematically learning dynamic cardiac motion from echocardiogram videos, extracting hemodynamic information, and learning metadata such as gender and age.
% MePH estimates mPAP and PVR without reliance on conventional echocardiographic parameters, by integrating multi-view imaging and metadata. 
% As illustrated in Fig.~\ref{fig:cam_visual}, 
Compared to expert echocardiographers' assessments via echocardiography, \methodname{} significantly reduced the MAE for mPAP and PVR assessments by 49.73\% and 43.81\%, respectively, achieving an MAE of 5.099 ± 0.107 for mPAP and 1.571 ± 0.043 for PVR relative to the invasive RHC approach. In eight independent external hospitals, \methodname{} demonstrated an MAE of 3.147 ± 0.072 for PVR assessment.
Moreover, \methodname{} achieved an area under the curve (AUC) of 0.921, outperforming echocardiographers, who achieved an AUC of 0.842, in accurately predicting PH severity. 
Our MePH model can effectively identify clinically relevant regions in echocardiogram videos, such as the interventricular septum and pulmonary artery wall, by learning key visual features. This capability allows the model to detect early structural changes associated with increased pressure and resistance, enabling a non-invasive assessment of PH severity and supporting treatment planning.
This model seamlessly learns cardiac motion and structural characteristics from echocardiogram videos, critical hemodynamic information from Doppler images, and patient demographic data. By effectively aligning and integrating these diverse modalities, MePH promotes a comprehensive understanding of PH patterns, enhancing analytical capabilities for detecting and assessing PH severity.

% Furthermore, \methodname{} improved the accuracy of PVR assessment by 43.81\%, reaching an MAE of 1.571 ± 0.043 compared to the invasive RHC approach.

\if 1 
Compared to the existing method used by Echocardiograhers based on echocardiography, \methodname{} significantly reduced the MAE of mPAP and PVR assessments by 49.73\% and 43.81\%, respectively, achieving an MAE of 5.099 ± 0.107 for mPAP and 1.571 ± 0.043 for PVR relative to the invasive RHC approach. In eight independent external hospitals, \methodname{} achieved an MAE of 3.147 ± 0.072 for PVR assessment.
Furthermore, \methodname{} achieved an AUC of 0.921, surpassing Echocardiograhers, who had an AUC of 0.626, in accurately predicting PH severity, whether mild or severe.
Therefore, our MePH can be utilized to predict treatment efficacy for PH patients. A prospective study demonstrated that MePH shows promising results in predicting treatment efficacy for patients.
\fi 

  \begin{figure}[!]
    \centering
    \includegraphics[width=0.999\linewidth]{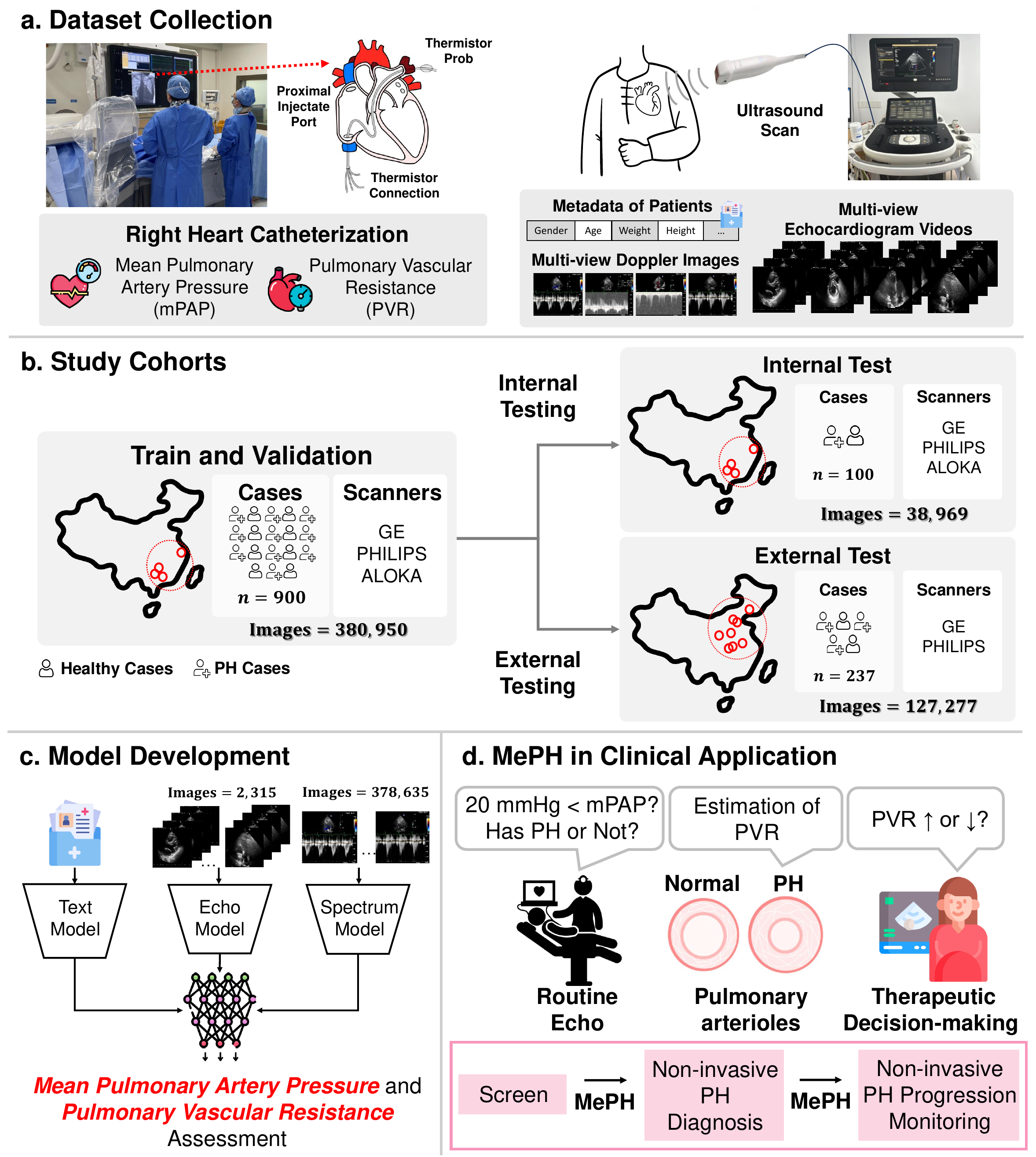}
    \vspace{-10pt}
    \caption{\textbf{Overview of our study.} \textbf{a, Dataset collection} shows that each patient underwent two types of examinations: 1. Echocardiography performed by echocardiographers, which includes echocardiogram videos from four views, spectral images from four views, and metadata; 2. RHC, which serves as the ``gold standard'' for accurately obtaining mPAP and PVR.
    \textbf{b, Study cohorts} presents the data statistics of our multi-center PH dataset, which includes a total of 1,237 patients from 12 medical centers. \textbf{c, Model development} shows that our MePH model was trained on a collected multi-view, multi-modality dataset to learn dynamic cardiac motion from echocardiogram videos, extract hemodynamic information, and incorporate metadata such as gender and age.
    \textbf{d, Real world clinical application} of our MePH shows its capability to predict the severity of PH and assess treatment efficacy, ultimately enhancing overall survival for PH patients. }
    \label{fig:overview}
  \end{figure}
\section*{Results}
\subsection*{Datasets Collection and Study Cohorts}
% As shown in Figure \ref{fig:overview}, the overview of our work comprises four parts.
As shown in Figure \ref{fig:overview}, our work consists of four main parts.
First, we established a multi-center dataset for model training and evaluation. We collected a total of 1,000 cases of patients with PH and non-PH conditions from four centers in China: Guangdong Provincial People’s Hospital (denoted as A), The First Affiliated Hospital of Guangzhou Medical University (denoted as B), Tongji University Affiliated Shanghai Pulmonary Hospital (denoted as C), and Fuwai Hospital Chinese Academy of Medical Sciences (denoted as D).
For external validation, we collected 237 cases from eight medical centers in China: Nanjing First Hospital (denoted as E), Tianjin Medical University General Hospital (denoted as F), The First Affiliated Hospital of Zhengzhou University (denoted as G), The First Affiliated Hospital of Xi'an Jiaotong University (denoted as H), Wuhan Asia Heart Hospital (denoted as I), Beijing Anzhen Hospital (denoted as J), Tongji Hospital, Tongji Medical College of HUST (denoted as K), and General Hospital of Northern Theater Command of the Chinese People's Liberation Army (denoted as L). Further details can be found in Supplementary Table~\ref{tab:ph_etiology}.

% As shown in Supplementary Table~\ref{table:dataset_statics}, all cases in clinically stable condition, aged 14 to
% 77 years, underwent RHC to measure precise mPAP$_\text{RHC}$ and PVR$_\text{RHC}$. Within 48 hours, the experienced echocardiographers scanned echocardiogram videos from four views, which are the Apical Four-chamber view (A4C), Parasternal Pulmonary Artery Long Axis view (PALA), Parasternal Short Axis-Papillary Muscle view (PSAX), and Parasternal Left Ventricular Long Axis view (PLAX). Additionally, 4 views of Doppler images from the Right ventricular outflow tract antegrade blood flow (RVOT), Pulmonary Regurgitation (PR), Tricuspid Regurgitation (TR), and Pulmonary valve anterior flow spectrum (PV), as well as their metadata such as 
% age, gender, height and weight et al., are collected synchronously.
% In our experiment, the cases from centers A, B, C, D ($n=1,000$) and E, F, ..., L ($n=237$) were divided into internal and external datasets, respectively. The internal dataset was randomly separated into 800 cases for training, 100 cases for validation and 100 cases for testing. 

As shown in Supplementary Table~\ref{table:dataset_statics}, all cases involved clinically stable individuals aged 14 to 77 years who underwent RHC to measure mPAP$_\text{RHC}$. Within 48 hours, experienced echocardiographers scanned echocardiogram videos from four views: the Apical Four-Chamber view (A4C), Parasternal Pulmonary Artery Long Axis view (PALA), Parasternal Short Axis-Papillary Muscle view (PSAX), and Parasternal Left Ventricular Long Axis view (PLAX). Moreover, Doppler images were collected from four views of the Right Ventricular Outflow Tract antegrade blood flow (RVOT), Pulmonary Regurgitation (PR), Tricuspid Regurgitation (TR), and Pulmonary Valve anterior flow spectrum (PV), along with metadata such as age, gender, height, and weight. In our experiment, cases from centers A, B, C, and D (n=1,000) were designated as the internal dataset, while those from centers E, F, ..., and L (n=237) formed the external dataset. The internal dataset was randomly divided into 800 cases for training, 100 cases for validation, and 100 cases for testing. We trained the MePH model and assessed its classification performance using mPAP within the internal dataset, as shown in Figure~\ref{fig:overview}c. Subsequently, we evaluated the model's generalizability on external data. To determine the effectiveness of MePH in assessing the severity of PH, we compared the predicted PVR from our MePH with those obtained from RHC, examining the concordance of changes in PVR between the two methods.

We also conducted a prospective study to demonstrate the effectiveness of our MePH for non-invasive assessment of PH. As shown in Supplementary Figure~\ref{fig:inclusion_exclusion_cascade}, our external dataset includes a total of 110 patients who underwent RHC before treatment and a follow-up RHC after 6 months of percutaneous pulmonary artery denervation. 
By utilizing only echocardiography, we will show the effectiveness of our MePH for assessing mPAP and PVR, highlighting its utility in monitoring the severity of PH.

% Our constructed study cohort is able to serve multiple clinical applications in real scenarios, including PH detection, severity prediction, and PH follow-up evaluation. This analysis involved a cohort of 110 cases, comparing measurements taken during the initial assessment and those obtained three months post-RHC.

% Cardiovascular disease specialists, including both junior and senior professionals, were involved in verifying the effectiveness of MePH in the clinical environment. These specialists first provided manual assessments of mPAP$_\text{Echo}$ and PVR$_\text{Echo}$ for PH by echocardiography. Then, they re-evaluated assessments with feedback from the MePH. Finally, we validated the effectiveness and compatibility of the MePH by analyzing its output patterns and comparing them to the diagnostic tendencies observed in clinical practice. A detailed description of the experimental procedures and supplementary materials is provided in our Methods section.

%
  \begin{figure}[t!]
    \centering
    \makebox[\textwidth][c]{\includegraphics[width=1.23\textwidth]{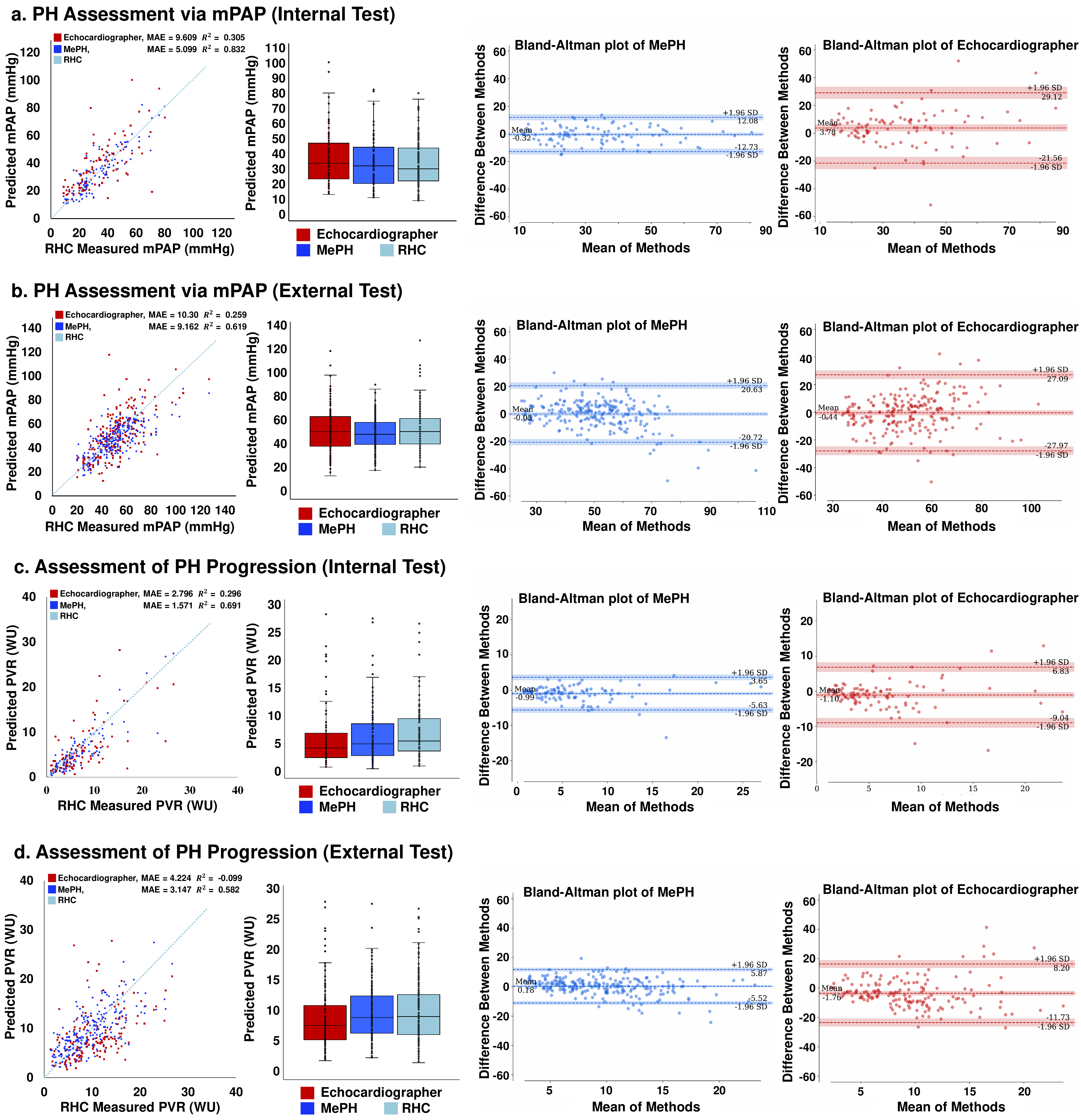}}
    \caption{\textbf{Comparison of PH recognition and PVR prediction results for MePH, echocardiographers, and RHC in both internal and external tests.} \textbf{a,} Results of PH recognition from the internal test set across four medical centers (A, B, C, D) with a test sample size $n=100$. \textbf{b,} Results of PH progression assessment from the external test collected from eight medical centers (E, F, ..., L) with a test sample size of $n=237$. \textbf{c,} Results of PH recognition from the internal test set across four medical centers (A, B, C, D) with a test sample size $n = 100$.  \textbf{d,} Results of PH progression assessment from the external test were collected from eight medical centers (E, F, ..., L) with a test sample size of $n=237$. The Bland-Altman plots are presented based on the estimated mPAP and PVR compared to the ground-truth mPAP and PVR measured by RHC, illustrating the limits of agreement (dotted lines) ranging from -1.96 to +1.96 standard deviations. }
    \label{fig:diagnosis_result_mpap}
  \end{figure}
  \begin{figure}[!]
    \centering
    \makebox[\textwidth][c]{\includegraphics[width=1.15\textwidth]{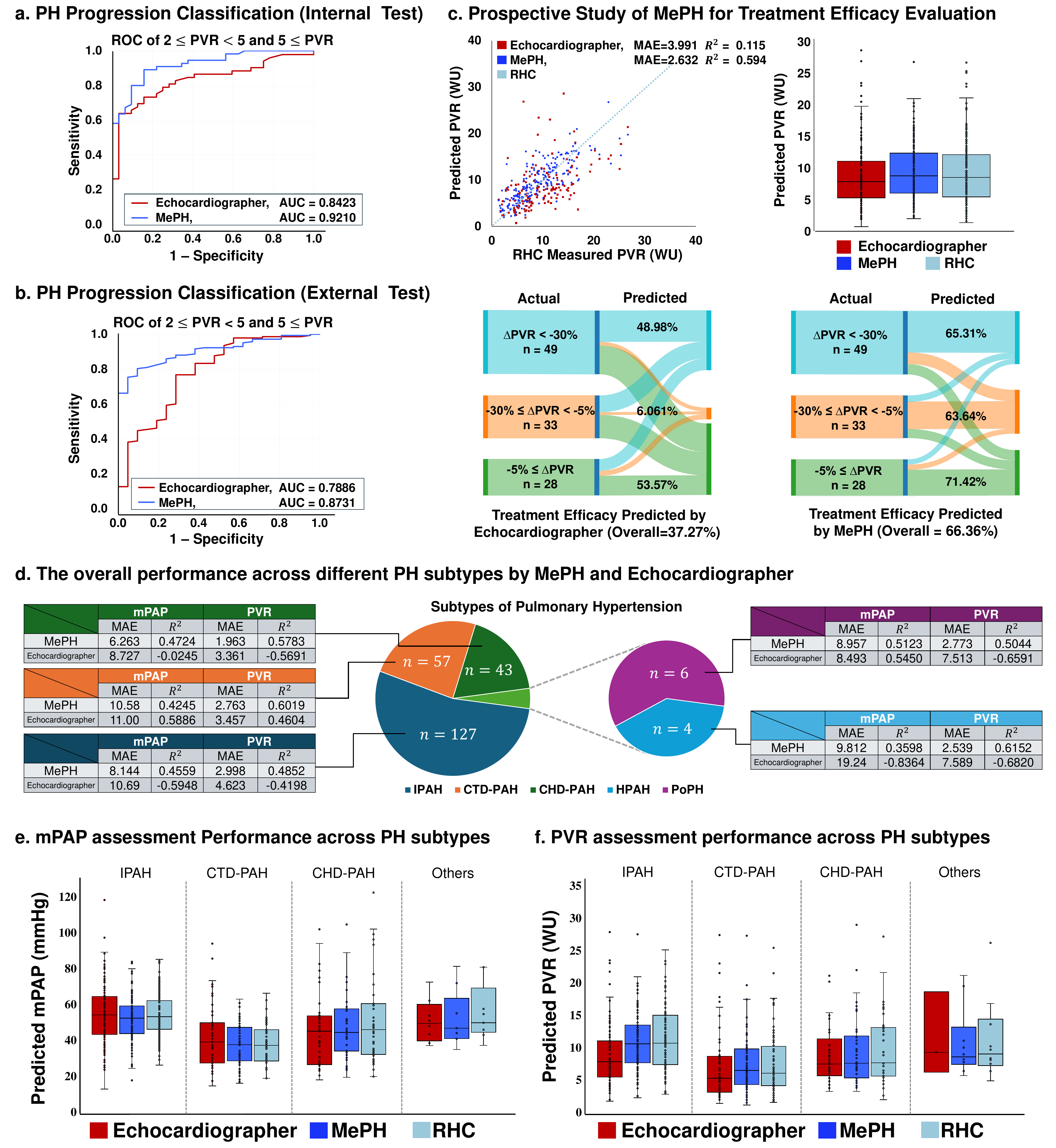}}
    \caption{ 
    \textbf{The PH progression assessment efficacy, the prospective study of treatment efficacy, and the performance comparison across different subtypes of PH.} 
    \textbf{a,} The PH progression assessment efficacy in the internal test set. \textbf{b,} The PH progression assessment efficacy in the external test set \textbf{c,} A prospective study to show the effectiveness of MePH for evaluating treatment efficacy. The predicted ($\Delta\text{PVR}$) is calculated for the same patients by comparing the initial $\text{PVR}^\text{1st}$ accessed by RHC and the follow-up $\text{PVR}^\text{2nd}$ accessed by the echocardiographer and our MePH, respectively ($\Delta\text{PVR}=\text{PVR}^\text{2nd}-\text{PVR}^\text{1st}$). The actual ($\Delta\text{PVR}$) is calculated by RHC over the two stages.
    \textbf{d,} The overall performance across different PH subtypes. All results in tables of each PH subtype were reported by mean absolute error (MAE) and the coefficient of determination $R^2$ for PH classification based on mPAP (mmHg) and PH progression assessment based on PVR (WU). \textbf{e,} The performance of mPAP assessment in IPAH, CTD-PAH, CHD-PAH and others (HPAH and PoPH). \textbf{f,} The performance of PVR assessment in IPAH, CTD-PAH, CHD-PAH and others (HPAH and PoPH). All results were reported in the external dataset. For the full names of these Pulmonary Hypertension subtypes, please see our Supplementary Table~\ref{tab:ph_etiology}.
}
    \label{fig:diagnosis_result_pvr}
  \end{figure}

\subsection*{Evaluation of MePH for mPAP Assessment}
Figure~\ref{fig:diagnosis_result_mpap}a shows the results of MAE for mPAP evaluated by echocardiographers and our MePH, compared to the ground-truth RHC. The findings show that our MePH (MAE = 5.099$\pm$0.107) exhibits a significantly smaller error than echocardiographers (MAE = 9.609) and demonstrates fewer outliers.
In the $R^2$ analysis, MePH achieves an $R^2$ of 0.832$\pm$0.037, significantly surpassing echocardiographers, who have an $R^2$ of 0.305 using the first-line diagnostic method~\cite{abbas2013noninvasive}. This indicates that MePH is highly correlated with the real measured mPAP and is statistically significant. 
The first-line diagnostic method used by echocardiographers typically relies on only two echocardiographic parameters, such as estimated Right Atrial Pressure (eRAP) and Tricuspid Regurgitation Velocity 
(TRV), to estimate mPAP$_\text{Echo}$.
However, as illustrated in the scatter and box plots in Figure~\ref{fig:diagnosis_result_mpap}a, echocardiographers tend to underestimate mPAP. This observation highlights the limitations of relying on isolated and restricted parameters for mPAP estimation, which can lead to inconsistencies in accuracy and potential misclassification of PH. 
The results further show that our MePH can learn various representations of PH through multiple modalities, including echocardiogram videos, Doppler images, and metadata. This multifaceted approach enhances the learning process related to PH through echocardiography, ultimately leading to a more accurate evaluation by MePH.

\subsection*{Evaluation of MePH in PVR Assessment}

According to the guidelines established by Humbert et al.~\cite{humbert20222022}, the severity and progression of PH can be assessed through the evaluation of PVR. In this section, we compare the PVR predictions of our MePH method with those of echocardiographers, alongside the results obtained via RHC. As shown in the scatter and box plots in Figure~\ref{fig:diagnosis_result_mpap}c, the MePH method achieves a significantly lower MAE of 1.571±0.043, outperforming the echocardiographer assessments, which exhibit an MAE of 2.796. These results indicate that the echocardiographer method—based on total vascular resistance (TVR) and tricuspid valve inflow (TVI$_\text{RVOT}$)—tends to overestimate PVR compared to RHC measurements. This overestimation can lead to misinterpretation of PH severity and may result in inappropriate therapeutic decisions.

Furthermore, MePH shows superior performance in predicting the severity of PH, as shown in Figure~\ref{fig:diagnosis_result_pvr}a.
The AUC for MePH is 0.9210 (95\% CI 0.915 – 0.925) of classifying cases from 2 $\leq$ PVR < 5 and 5 $\leq$ PVR, significantly surpassing the AUC reported by echocardiographers of 0.8423. For statistically significant benefits, our MePH reached an $R^2$ of 0.691 $\pm$ 0.024, surpassing the $R^2$ reported by echocardiographers of 0.305, indicating our MePH can accurately predict the PH progression. This result shows that echocardiographers relying solely on echocardiogram-derived parameters, such as eRAP and TRV, are insufficient for accurately assessing PH. In contrast, our MePH leverages multi-view, multi-modal information in echocardiography to achieve a more precise evaluation.

\subsection*{Evaluation of MePH for PH and Non-PH Classification}

The precise assessment of mPAP facilitates a more accurate classification of PH without the need for invasive procedures. In this section, we compare the effectiveness of MePH and echocardiographers in classifying PH and non-PH cases and detecting PVR abnormality (cases with mPAP < 20 and PVR < 2 defined as non-PH\cite{humbert20222022}).
As shown in Supplementary Figure~\ref{fig:classification_result}, the MePH model achieves an AUC of 0.9178 (95\% confidence interval [CI] 0.913–0.921) based on mPAP criteria for PH classification and 0.9514 (95\% CI 0.948–0.954) based on PVR criteria for PH classification. 
In contrast, echocardiographers using the existing first-line diagnostic methods, eRAP and TRV, achieve AUCs of 0.9122 and 0.7953, respectively.
Our MePH model outperforms the echocardiographers by a clear margin, showing improvements of 0.0056 and 0.1561 in AUC for PH and non-PH classification based on mPAP and PVR criteria, respectively. 
This demonstrates the potential of MePH for rapid and accurate screening of PH. 

\subsection*{A Prospective Study of MePH for Treatment Efficacy Evaluation}

We conducted a prospective study to evaluate the effectiveness of our MePH in assessing treatment efficacy for patients with PH. A total of 110 patients from independent external hospitals were included, all of whom underwent RHC prior to treatment and a follow-up RHC six months after percutaneous pulmonary artery denervation.
The change in $\Delta\text{PVR}$ was calculated as the difference between the initially assessed $\text{PVR}^\text{1st}$ and the follow-up $\text{PVR}^\text{2nd}$. $\Delta\text{PVR}$ was categorized into three groups: $\Delta\text{PVR} < -30\%$, $-30\% \leq \Delta\text{PVR} < -5\%$, and $-5\% \leq \Delta\text{PVR}$, indicating varying levels of therapeutic effectiveness.

Our MePH relies exclusively on echocardiography conducted within 24 hours of the follow-up RHC, which occurs six months after treatment, to predict PVR in post-treatment patients. By subtracting the PVR obtained from RHC prior to treatment, we can calculate the $\Delta\text{PVR}$ to evaluate treatment efficacy.
As shown in the scatter and box plots in Figure~\ref{fig:diagnosis_result_pvr}c, our MePH achieved an MAE of 2.632 ± 0.06, compared to an MAE of 3.689 obtained by the echocardiographer in estimating the PVR after treatment.  

The Sankey diagram further illustrates patient-level information from the MePH model in predicting changes in PVR, contrasting with the estimates provided by echocardiographers.
The results revealed that the echocardiographer's performance in evaluating PVR changes was unsatisfactory, with accuracy rates of only 48.98\%, 6.061\%, and 53.57\% for detecting changes in each $\Delta\text{PVR}$ category, respectively. In comparison, our MePH significantly enhanced predictive performance, yielding accuracy rates of 65.31\% (an increase of 16.33\%), 63.64\% (an increase of 57.58\%), and 71.42\% (an increase of 17.85\%). These findings indicate that MePH can serve as a highly accurate non-invasive measure for monitoring treatment efficacy, reducing the need for invasive RHC for monitoring purposes. Our MePH offers a more accessible method for evaluating treatment efficacy during follow-up assessments, enabling echocardiographers to make better-informed treatment decisions.

\subsection*{Generalization to Unseen External Hospitals}
We evaluate the generalizability of MePH by selecting the model that performs best on the internal validation set. We then evaluate its performance on an independent external dataset, which comprises data collected from eight different hospitals.
For PH classification in the external dataset ($n=237$), 
For both mPAP and PVR assessment, Figures~\ref{fig:diagnosis_result_mpap}b and~\ref{fig:diagnosis_result_mpap}d show a slight degradation of MePH performed in the external data. MePH achieved an MAE of 9.162$\pm$0.360 ($R^2$=0.619) in the external set, decreasing from MAE of 5.099 ($R^2$=0.832) for accessing mPAP in the internal set. The MAE increases by 1.576, and $R^2$ decreases by 0.109 when assessing PVR. 

%The AUC of mPAP declines by around 12\%, decreasing from 0.9506 in the internal set to 0.8231 in the external set. The MAE of mPAP and PVR decrease by 4.075 and 1.576, respectively, when adopting the trained MePH to external test set. 

We attribute the performance drop to the following reasons:
i) Device Differences: The data were collected from different devices, leading to variations in distribution and shifts in low-level features such as brightness, noise, and color.
ii) Differences in Echocardiogram Video Quality: The internal and external data were collected from medical centers in different regions of China. Variations in scanning parameters, patient demographics, and the skill levels of echocardiographers involved in data collection may have contributed to shifts in features within the training and external test cohorts.
iii) Differences in Patient Severity Distribution: As shown in Figures~\ref{fig:diagnosis_result_mpap} and Figure~\ref{fig:diagnosis_result_pvr}, the severity distribution of patients varies between training and external test sets. In the internal dataset, patients are primarily concentrated around an mPAP of 33 mmHg and a PVR of 4 WU, whereas these values shift to 52 mmHg and 11 WU in the external dataset. However, we note that the $R^2$ value decreases only marginally in the external dataset, which underscores several positive attributes of our approach.  Our MePH method not only excels with the internal dataset but also sustains practical and reliable performance on independent, unseen data. These findings further illustrate MePH's consistency and effectiveness across various datasets. This consistent performance on external data reinforces the robustness and broad applicability of our method. Hence, despite the performance shifting, our method can also demonstrate significant improvement in mPAP and PVR assessment compared to the echocardiographer who achieved MAE in mPAP assessment of 10.30 with $R^2$ is 0.259 and PVR assessment of 4.224 with $R^2$ is -0.099, respectively.
Furthermore, as shown in Figure~\ref{fig:diagnosis_result_pvr}b, when assessing PH progression by PVR at different levels, MePH achieves AUC values of 0.8731 (95\% CI 0.869–0.877) in classifying cases of $2\leq \text{PVR}<5$ and $5\leq\text{PVR}$, while echocardiographers have AUC values of 0.7886. The performance gaps of 0.0845 further reinforce the superior capability of MePH in assessing PH progression compared to echocardiographers.

% Furthermore, as shown in Figure~\ref{fig:diagnosis_result_pvr}, when assessing PH progression by PVR at different levels, 
% MePH achieves AUC values of 0.9061 (95\% CI 0.902–0.910) of $2\leq \text{PVR}\leq5$ and 0.8958 (95\% CI 0.892–0.899) and $2< \text{PVR}$, while echocardiographers have AUC values of 0.7560 and 0.7465, respectively.
% The performance gaps are 0.1501 and 0.1493, which reinforcing the superior position of MePH in PH progression assessment compared to echocardiographers.

% the performance gap is 0.1501 and 0.1493 of $2\leq \text{PVR}\leq5$ and $2< \text{PVR}$ between MePH with AUC of 0.9061 (95\% CI 0.0.902–0.910) and 0.8958 (95\% CI 0.892–0.899) and echocardiographers with AUC of 0.7560 and 0.7465, respectively, 
%
\begin{figure}[t!]
\centering
\makebox[\textwidth][c]{\includegraphics[width=1.07\textwidth]{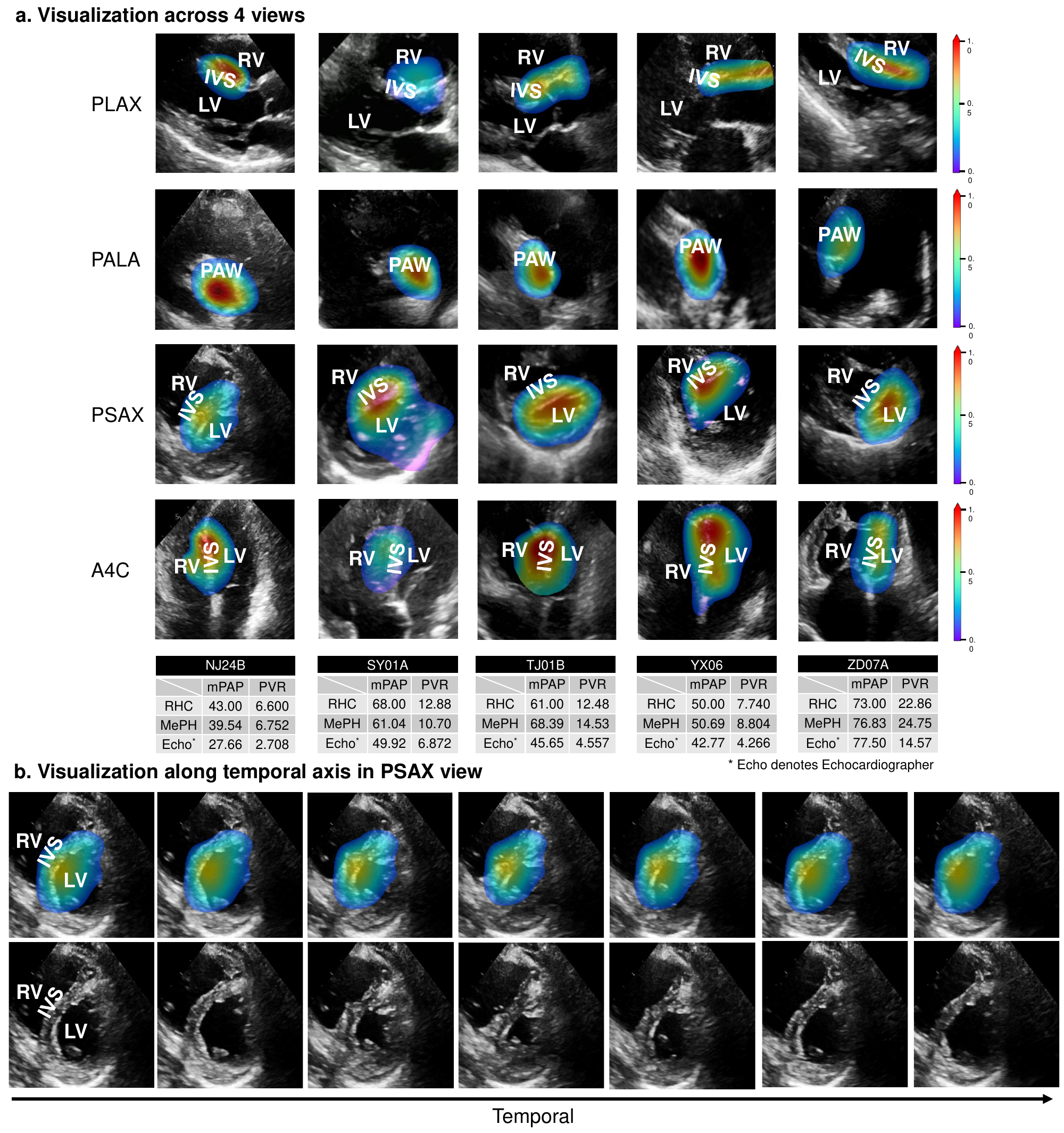}}
\vspace{-6pt}
\caption{\textbf{Illustration of the model decision with activation heatmap on different cases with PH.} The saliency map (heat map) was generated by EigenCAM~\cite{muhammad2020eigen} with no class discrimination, which is used to visualize which areas help the model assess the mPAP and PVR. The scale bar spans from 0 to 1, with a value of 1 representing the highest level of influence derived from the normalized EigenCAM value, while 0 indicates the lowest possible influence. \textbf{a,} The activation maps from four cardiac views can be used to observe cardiac motion. The highlighted active regions—the interventricular septum (IVS)—separate the right ventricle (RV) from the left ventricle (LV), as well as the pulmonary artery wall (PAW). The tables at the bottom present mPAP and PVR, measured by RHC, evaluated by MePH, and assessed by echocardiographers through echocardiography.
\textbf{b,} Images show the correlation of temporal and spatial information presented by cardiac structures in PSAX views. MePH highlights clinically relevant regions in echocardiogram frames, reflecting structural changes associated with elevated pressure and vascular resistance. These patterns correspond to established indirect signs of PH. In contrast to conventional estimation methods, MePH captures such features without requiring echocardiography signals, enabling parameter-free interpretation and contributing to the model’s superior performance in assessing mPAP and PVR.}
\label{fig:cam_visual}
\end{figure}
\subsection*{Generalization to Different Ultrasound Devices}

To further evaluate the generalization ability of MePH across different devices, we grouped the external data set scanned by the GE devices ($n = 164$) and PHILIPS ($n = 73$) and conducted experiments, respectively. As shown in Supplementary Figure~\ref{fig:cross_device_result}, a noticeable performance gap exists between devices, with the model derived from PHILIPS significantly outperforming that from GE in both mPAP and PVR assessments.
This outcome can primarily be attributed to the imbalanced representation of different echocardiography devices (e.g., PHILIPS, GE, ALOKA) in our training data.
As indicated in Supplementary Table~\ref{table:dataset_statics}, the internal dataset comprises 72.7\% from PHILIPS, 15.5\% from GE, and 11.8\% from ALOKA, whereas the external data are primarily concentrated on GE, with a ratio of around 2:1 compared to PHILIPS.
This imbalanced data composition results in the model achieving higher accuracy with PHILIPS data compared to GE in the internal set, and it also contributes to domain-shifting issues between the internal and external test sets.
Despite the existing domain gap, MePH achieves impressive performance in mPAP assessment, with MAE values of  9.836$\pm$0.121 for GE and 7.594$\pm$0.082 for PHILIPS. This outperforms echocardiographers by 10.42\% (MAE = 10.98) for GE and 20.61\% (MAE = 9.565) for PHILIPS.
In the PVR assessment, MePH achieves MAE values of 3.353$\pm$0.076 for GE and 2.449$\pm$0.063 for PHILIPS, significantly surpassing echocardiographers by 16.38\% (MAE = 4.010) and 47.78\% (MAE = 4.690), respectively.
These results emphasize that, despite minor variations in performance across different devices, MePH consistently outperforms echocardiographers in assessing mPAP and PVR. This leads to more accurate evaluations and empowers echocardiographers to make better-informed treatment decisions.

\subsection*{Generalization to Different Pulmonary Hypertension Subtypes}
We assess the generalizability of MePH across diverse PH subtypes. As detailed in Supplementary Table~\ref{tab:ph_etiology}, the PH cases of the external dataset can be categorized into IPAH (n=127), HPAH (n=4), CTD-PAH (n=43), CHD-PAH (n=57), and PoPH (n=6), with IPAH, CTD-PAH, and CHD-PAH constituting the majority. The internal dataset predominantly comprises CTEPH (n=385), IPAH (n=199), and CHD-PAH (n=211). Consequently, this allows us to examine MePH's adaptability to rare or previously unseen PH subtypes and compare its efficacy against traditional echocardiographic parameter-based diagnoses by echocardiographers.

The comparative performance of MePH and echocardiographers across various PH subtypes are summarized in Figure~\ref{fig:diagnosis_result_pvr}d. This figure demonstrates that MePH achieves superior accuracy in terms of MAE and $R^2$ for both mPAP and PVR relative to echocardiographers. For instance, in the assessment of mPAP among IPAH and CHD-PAH patients—major categories in both datasets—MePH records an MAE of 8.144$\pm$0.143 mmHg and an $R^2$ of 0.4559$\pm$0.048 for IPAH, while echocardiographers report an MAE of 10.69 mmHg and an $R^2$ of -0.5948. In CHD-PAH, MePH achieves an MAE of 6.263$\pm$0.109 mmHg with an $R^2$ of 0.4724$\pm$0.044, surpassing echocardiographers' MAE of 8.727 mmHg with an $R^2$ of -0.0245 with a clear margin. Additionally, in PVR assessments for these two subtypes, IPAH and CHD-PAH, MePH outperforms echocardiographers with respective MAEs of 2.998 WU and 1.963 WU versus 4.623 WU and 3.361 WU.

Regarding PH subtypes not extensively covered in the internal dataset, specifically CTD-PAH and HPAH, our analysis reveals that MePH exhibits superior performance in the assessment of mPAP and PVR compared to echocardiographers. For PoPH, MePH showed a slight decline in estimated mPAP but a significant improvement in PVR evaluation. This consistent superiority underscores the robustness and reliability of MePH in addressing a wide spectrum of PH subtypes. Figure~\ref{fig:diagnosis_result_pvr}e highlights the performance of mPAP assessment across different PH subtypes, demonstrating MePH's consistent advantage over echocardiographers, especially in IPAH, CTD-PAH, and CHD-PAH. Similarly, Figure~\ref{fig:diagnosis_result_pvr}f illustrates MePH's enhanced predictive accuracy for PVR assessments across all subtypes, paralleling its mPAP performance. These outcomes collectively affirm MePH's potential as a reliable tool for clinical applications involving diverse PH subtypes.
\begin{figure}[t]
\centering
\makebox[\textwidth][c]{\includegraphics[width=1.3\textwidth]{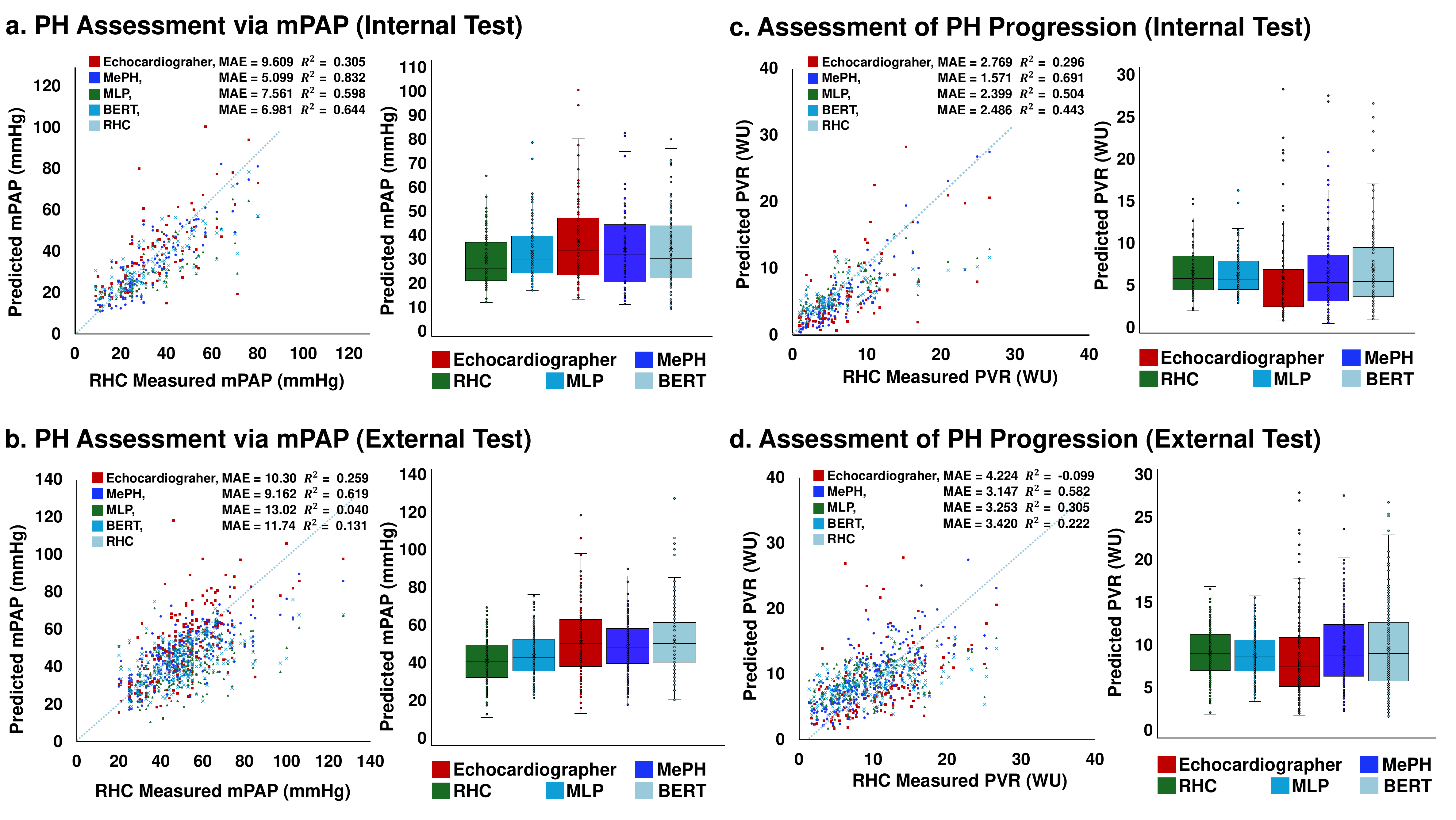}}
\caption{\textbf{Comparison of PH recognition and PVR prediction with vision-language based method and echocardiography parameter-based methods}. \textbf{a,} Results of PH recognition from the internal test set across four medical centers (A, B, C, D) with a test sample size $n = 100$. \textbf{b,} Results of PH progression assessment from the external test collected from eight medical centers (E, F, ..., L) with a test sample size of $n = 237$. \textbf{c,} Results of PH recognition from the internal test set across four medical centers (A, B, C, D) with a test sample size $n = 100$. \textbf{d,} Results of PH progression assessment from the external test collected from eight medical centers (E, F, ..., L) with a test sample size of $n = 237$. All results are reported by metrics MAE and $R^2$, PH classification is assessed by mPAP and PH progression is assessed by PVR.}
\label{fig:MLP_BERT_result}
\end{figure}

\subsection*{Human-Centered Evaluation of Model Explainability}
To evaluate the explainability of MePH, we utilized EigenCAM~\cite{muhammad2020eigen} without the class discrimination module to highlight the regions of the echocardiogram video that assist MePH in assessing mPAP and PVR. We then invited 4 expert echocardiographers to analyze the highlighted results from the echocardiogram videos and determine if they align with clinical guidelines.

After evaluating 237 cases from all external test set, the echocardiographers reached the following consensus and conclusions: 
% 首先，所有的医生对识别的结构达成了共识，模型识别的主要结构包含这三个。接下来对三个结构逐个进行解释。
The main regions identified by our model include interventricular septum (IVS), pulmonary artery wall (PAW) and left ventricle (LV), as shown in Figure \ref{fig:cam_visual}. These identified regions are consistent with the indirect signs of pulmonary hypertension as outlined in clinical guidelines~\cite{humbert20222022}. 
% 对最重要、出场频率最高的结构进行解释。第一，IVS可以反映左右心室压力差；第二，IVS形变发生的时相也反映了压力负荷或是容量负荷的信息。（所以IVS这个结构是合理的）
Specifically, (1) the interventricular septum (IVS) is identified in three echocardiographic views: PLAX, PSAX, and A4C. It serves as a key structure for predicting pressure and resistance. Changes in the pressure differential between LV and RV can lead to the deformation of the IVS. Moreover, different types of hemodynamic loading—specifically pressure or volume overload—can cause IVS deformation at various phases of the cardiac cycle~\cite{yamasaki2016clinical}, as shown in Figure~\ref{fig:cam_visual}b. 
% 对第二个结构，即肺动脉壁进行解释。肺动脉壁提供了血管顺应性(compliance)的信息，顺应性下降，阻力会升高。（所以这个结构也是合理的）
(2) The PAW is highlighted in every PALA view, which may offer crucial information about vascular compliance.
In PH, the pulmonary vasculature becomes pathologically non-compliant as a result of vascular fibrosis and stiffening~\cite{thenappan2018pulmonary}. A decrease in vascular compliance manifests as an increase in PVR. 
% 对第三个结构，即左室进行解释。左室的形态反映了压力和阻力的升高情况，即可以提供压力和阻力的信息。（所以这个结构也是合理的）
(3) LV chamber is also recognized in the PSAX view, which aligns with existing knowledge. The morphology of the LV provides information on the degree of pressure and resistance elevation~\cite{matsumura2021interventricular,critser2020cardiovascular,dellegrottaglie2007pulmonary}. Therefore, the regions highlighted by the model are considered reasonable based on the understanding of echocardiographers.

% 总结，表明以上我提到的所有结构，在医生看来，都是合理的、有诊断价值的。

These findings highlight the necessity for further research to clarify the complex interactions between left and right heart function in contributing to elevated pulmonary pressure and resistance. While echocardiographers mainly rely on established Doppler echocardiography hemodynamic parameters rather than visual features for clinical diagnosis, their interpretations can be significantly affected by variations in scanning quality. These variations can impact the measured parameters, thereby limiting the accuracy of assessments related to cardiac structural and motion abnormalities. Future investigations should prioritize longitudinal studies that monitor changes in cardiac morphology and function over time, correlating these changes with clinical outcomes in PH patients~\cite{mouratoglou2020duration}.
In conclusion, the insights from Figure~\ref{fig:cam_visual} show the intricate relationship between resistance and cardiac structures, underscoring the importance of a comprehensive evaluation of both the right and left heart in PH. These findings advocate for a shift in focus that recognizes the significant interplay between the two sides of the heart, ultimately enhancing echocardiographers' understanding and management of this complex condition.

\subsection*{MePH Outperforms Parameter-based Echocardiographic Assessment of Pulmonary Hypertension}

We conducted a comprehensive comparative evaluation against parameter-based learning models (MLP, BERT~\cite{devlin2019bert}, DistilBERT~\cite{sanh2019distilbert}) and expert echocardiographers to validate the diagnostic strength of MePH. As shown in Figure~\ref{fig:MLP_BERT_result} and detailed in Supplementary Table~\ref{tab:performance_across_four_methods}, experiments across both internal and external cohorts consistently demonstrate that MePH substantially outperforms all baselines in estimating mPAP and PVR. For the mPAP estimation on the internal dataset, MePH achieved the lowest mean absolute error (MAE) of 5.099 mmHg and the highest coefficient of determination ($R^2$ = 0.832), outperforming MLP (MAE = 7.561, $R^2$ = 0.598) and BERT-base (MAE = 6.981, $R^2$ = 0.644). Notably, echocardiographers' assessments yielded a significantly higher MAE of 9.609 and a lower $R^2$ of 0.305. These trends were consistent in the external dataset, where MePH maintained a strong performance (MAE = 9.162, $R^2$ = 0.619), while MLP and BERT experienced considerable performance degradation (MAE = 13.02 and 11.74, respectively). A similar but more pronounced pattern emerged in the estimation of PVR, which serves as a more direct signal of PH progression. Internally, MePH attained an MAE of 1.571 WU and $R^2$ = 0.691, significantly outperforming MLP (MAE = 2.399) and BERT (MAE = 2.486). Externally, MePH continued to lead with an MAE of 3.147 WU and $R^2$ = 0.582, whereas BERT and MLP reported lower accuracies and weaker correlations with ground truth. Among the parameter-based baselines, BERT-base achieved the best overall accuracy but incurred substantial computational cost, while DistilBERT offered a more efficient trade-off with moderate performance. In contrast, MLP, despite being extremely lightweight, consistently underperformed due to limited representational capacity. These observations highlight that increasing model size does not guarantee significantly better outcomes, and text-only architectures face inherent limitations in capturing complex clinical patterns.

In addition to accuracy metrics illustrated in Supplementary Table~\ref{tab:performance_across_four_methods}, MePH also demonstrated competitive computational characteristics. Despite its larger parameter size (194.3M) and higher computational load (904.98 GFLOPS), the inference time (45.21 ms) remains within an acceptable range for clinical deployment. In contrast, MLP and BERT-base, although lighter, achieved inferior predictive performance (e.g., RMSE of 17.11 and 16.27 for mPAP in the external dataset, respectively). This consistent superiority is attributed to MePH’s vision-centric, multi-modal design that fuses echocardiogram videos, Doppler images, and patient metadata. Unlike MLP and BERT, which rely solely on structured or textual data, MePH captures spatial-temporal patterns and clinically relevant features from echocardiographic inputs. This not only enables superior accuracy and generalizability but also promotes interpretability and trustworthiness, establishing MePH as a robust and clinically valuable tool for non-invasive assessment of pulmonary hypertension.

\section*{Discussion}
%% overview of paper contributions 
In this paper, we propose a multi-view, multi-modal vision-language model, MePH, to accurately evaluate PH progression using non-invasive echocardiography. Our goal is to provide PH patients with a non-invasive method for monitoring their disease prognosis, rather than relying solely on invasive RHC.
With MePH, the hemodynamic parameters of PH—mPAP and PVR—which are crucial for accurately assessing the severity of PH, can be reliably evaluated during routine echocardiography examinations throughout the entire diagnostic process, regardless of PH subtypes and scanning techniques.
Importantly, MePH consistently outperforms conventional echocardiographic assessments (i.e., echocardiographers' results in this manuscript) in both mPAP and PVR across eight external centers and various ultrasound devices.
Moreover, human-centered evaluations of our model predictions further strengthen echocardiographers' trust in our model and enhance their understanding of PH through echocardiography.

%% impact to clinical practices. 
Our work offers a valuable contribution to current clinical practices. While echocardiographers can identify PH and Non-PH conditions through echocardiography, they often face challenges in accurately predicting critical measurements such as mPAP and PVR. The reliance on RHC for these measurements is invasive and not recommended as a primary screening method, limiting its effectiveness for monitoring PH progression and evaluating treatment outcomes. It is important to clarify that we do not advocate for the replacement of RHC, which remains the gold standard. Rather, our model serves as an adjunctive tool, providing a novel, non-invasive, and accurate approach to monitoring PH progression in patients. As illustrated in Figure~\ref{fig:overview}d, by integrating this model into the diagnostic workflow, we aim to enhance the precision and efficiency of PH management, enabling earlier interventions and more personalized treatment decisions.

Our MePH can benefit current clinical practice in three key ways.
First, MePH is a highly accurate, non-invasive method for evaluating the severity of PH, surpassing traditional echocardiographic indicators such as TRV by providing a more precise assessment of mPAP and PVR.
An accurate evaluation method for PH is essential for timely assessments of severity and the development of effective treatment plans. Current clinical practices have several limitations. While TRV is a critical echocardiographic indicator for classifying PH, there are cases where TR may go undetected. However, the absence of identifiable regurgitation signals does not imply that PH is absent, which diminishes diagnostic sensitivity.
Furthermore, although TRV-derived pressure shows a strong correlation with RHC (with a high correlation coefficient R), it lacks good consistency (wide limits of agreement)~\cite{d2022echocardiographic}. This results in significant discrepancies when applied to individuals, indicating potential inaccuracies.
Therefore, the latest guidelines recommend assigning an echocardiographic probability of PH based on TRV and other suggestive echocardiographic signs~\cite{humbert20222022}. However, this evaluation is often time-consuming, subjective, and reliant on the examiner's experience, making accurate and real-time PH assessments challenging.
%Previous studies~\cite{diller2022framework} have shown that deep learning models can detect morphological features beyond mere RV dilation to classify PAH, a specific subtype of PH. 
In our study, we utilize four echocardiogram video views, four Doppler image views, and routine echocardiographic parameters to predict mPAP. This multimodal information comprehensively encompasses various indirect signs of PH related to the ventricles, pulmonary artery, and right atrium. In the absence of TRV, mPAP can be predicted by our MePH based on features learned from echocardiogram videos and Doppler images.
In summary, our MePH model can automate the PH probability assessment process as recommended in the guidelines~\cite{humbert20222022}. Importantly, MePH facilitates a direct evaluation of mPAP, allowing for definitive classification of PH rather than merely indicating its possibility~\cite{ragnarsdottir2024deep, sun2024chamber, zuwei2024automatic, tison2019automated, bird2020assessment, dubrock2024electrocardiogram, liu2024deep, diller2022framework}. Our model exhibits significantly narrower limits of agreement compared to conventional methods (see Figure~\ref{fig:diagnosis_result_mpap}a,b and Figure~\ref{fig:diagnosis_result_pvr}a,b), highlighting improved precision and enhancing the accuracy of applications to individual patients. These findings further reinforce our assertion that MePH is a TRV-independent, objective, and accurate diagnostic method.

Second, MePH is an effective non-invasive method for evaluating disease progression and predicting treatment efficacy. It is important to note that mPAP is a key parameter for screening and initial diagnosis of PH. However, mPAP is not directly related to the severity of the patient's condition or prognosis~\cite{humbert20222022}. Instead, PVR serves as a better indicator of disease severity and treatment response~\cite{humbert20222022,d2022echocardiographic,noordegraaf2019pathophysiology}.
While there is a TR-driven method from echocardiography for assessing PVR that combines PAP and flow measurements~\cite{naing2017non}, these parameters can vary in accuracy and are often impractical in many situations, such as in cases of severe tricuspid insufficiency, where the pressures in the RV and RA are nearly equalized.
Current studies have shown that improvements in right ventricle morphological parameters, such as RV end-diastolic area (RVEDA), left ventricular eccentricity index (LVEI), and RA area, correlate with treatment-induced decreases in PVR. However, these improvements were only observed when there was a significant change in PVR~\cite{badagliacca2018prognostic,d2020risk}.
In comparison to existing studies, our MePH can accurately predict PVR, regardless of whether the resistance is mild-to-moderate or severely elevated. The activation heatmap in Figure~\ref{fig:cam_visual} further illustrates that the model's prediction of PVR highlights specific regions that are understandable to echocardiographers.
The highlighted active regions emphasize the IVS and the PAW rather than the dimensions of the cardiac chambers. One possible explanation is that subtle remodeling in the myocardium may occur before measurable changes in chamber size.
Our MePH achieves enhanced sensitivity by identifying changes that occur prior to alterations in chamber size, allowing for accurate assessment of PVR throughout the entire diagnostic follow-up.

Third, MePH has the potential to optimize current workflows and reduce unnecessary RHC, leading to more precise treatment and management of PH patients. By accurately assessing mPAP and PVR, our MePH can facilitate early screening for PH and evaluate the pre-capillary component in underserved communities, thereby potentially improving the early diagnosis rate of the condition.
For suspected PH cases, a PVR higher than 5 WU may indicate a severe pre-capillary component, prompting clinicians to refer the patient to specialized PH centers for care. For confirmed cases, MePH provides a non-invasive PVR assessment based on echocardiography follow-ups every 3-6 months.
Once MePH detects significant disease progression, RHC will be indicated to confirm the results and adjust therapy. The satisfactory performance of our MePH in tracking disease progression was demonstrated in an independent PH cohort over a 6-month observation period, as shown in Figure~\ref{fig:diagnosis_result_pvr}c and Figure~\ref{fig:diagnosis_result_pvr}d. While hemodynamic assessment is crucial for guiding treatment, direct hemodynamic measurements via RHC cannot be performed on all patients or conducted as routine examinations due to their invasive nature.
Our MePH can help determine the timing of RHC, thereby reducing costs and the need for invasive procedures during the follow-up period.

%It can also improve patient recruitment and minimize churn since it is based on routine echocardiography examinations without incurring any additional burden on patients.

\if 1 
The MePH can be beneficial to several desirable applications. 
First, MePH provides a highly accurate, non-invasive method for evaluating the severity of PH, surpassing traditional echocardiographic indicators such as tricuspid regurgitation velocity (TRV) by delivering a more precise assessment of mPAP and PVR. An accurate evaluation method for PH is essential for timely severity assessments and the formulation of prompt treatment plans. Existing clinical practices have several limitations; although TRV remains a key echocardiographic indicator for classifying PH, the tricuspid regurgitation (TR) jet can be challenging for echocardiographers to detect, and the absence of a TR signal does not rule out the possibility of PH, thus reducing diagnostic sensitivity.
Moreover, there is a wide limit of agreement (\emph{i.e}. the individual data points are widely dispersed) between TRV-derived pressure and invasively measured pulmonary pressure, masked by the high correlation coefficient~\cite{d2022echocardiographic}. Thus, it is recommended to assign an echocardiographic probability of PH based on TRV and additional echocardiographic signs suggestive of PH in the latest guidelines~\cite{humbert20222022}. 
\xmli{I am confused about doctors' input.}
\fi

Despite the promising results achieved by our MePH, several limitations should be considered in the future.
First, our model requires high-quality multi-view echocardiogram videos and Doppler images, meaning that echocardiographers should record in a standardized manner and perform quality control to ensure all data aligns with the experimental configuration. Consequently, in practical applications, the input echocardiography must meet quality standards comparable to those we specified and adhere to the inclusion-exclusion criteria illustrated in Supplementary Figure~\ref{fig:inclusion_exclusion_cascade}.
Second, other modalities can also reflect cardiac function, such as electrocardiograms~\cite{tison2019automated,bird2020assessment,dubrock2024electrocardiogram,liu2024deep} and cardiac auscultation~\cite{dennis2010noninvasive,kaddoura2016acoustic}, which could potentially provide additional information for evaluating PH severity. While incorporating these modalities may enhance the accuracy of PH severity assessment, it would require significant effort to gather training data and could complicate testing scenarios by necessitating more modalities. This will be considered in our future work.

% The introduction of different modalities can help the model understand cardiovascular and hemodynamics comprehensively. We will also further this study by incorporating advanced cardiac measurement techniques and more efficient modal fusion methods to acquire more accurate assessments.

% In our future work, we will collect more data that cover more scanning angle of cardiac views. Moreover, we will built a quality control system to automatically process input data.

\if 1 
\section*{Discussion}

In this paper, we propose a multi-view multi-modal vision-language model, MePH, to classify PH and assess disease progression using multi-modal echocardiography. With MePH, the most important PH  hemodynamic parameters, mPAP and PVR, can be accurately assessed in routine echocardiography examinations during the entire diagnostic work-up, regardless of PH subtypes and scanning techniques. Furthermore, MePH outperforms the conventional echocardiographic assessments in both mPAP and PVR. Notably, we are able to validate our findings in an independent external PH cohort. Our model provides a real-time, efficient and no-invasive approach for accurate PH classification and progression assessment assistance compared to right heart catheterization, where our results demonstrate the potential of a new noninvasive method in classification and follow-up of PH. 

The MePH can be beneficial to several desirable applications. 
First, a method with high accuracy is crucial to enhancing the precision of the disease classification. MePH can promote accurate and timely PH classification, which is the foundation for effective clinical treatment. In comparison, though tricuspid regurgitation velocity (TRV) remains the key echocardiographic indicator for classifying PH, the tricuspid regurgitation (TR) jet can be undetectable, and the lack of TR signal does not exclude the probability of PH~\cite{o2018lack}. Moreover, there is a wide limit of agreement (\emph{i.e}. the individual data points are widely dispersed) between TRV-derived pressure and invasively measured pulmonary pressure, masked by the high correlation coefficient~\cite{d2022echocardiographic}. Thus, it is recommended to assign an echocardiographic probability of PH based on TRV and additional echocardiographic signs suggestive of PH in the latest guidelines~\cite{humbert20222022}. However, evaluating PH signs is time-consuming, subjective, and experience-dependent, where such uncertain estimation indicators can hardly provide accurate and real-time PH assessments. Previous study~\cite{diller2022framework} has shown that deep learning models can detect morphological features beyond merely right ventricle (RV) dilation to classify PAH, a specific subtype of PH. In our study, we utilize 4 echocardiogram video views, 4 Doppler image views, and echocardiography parameters routinely obtained from echocardiographic data to predict mPAP. This multi-modal information comprehensively covered a number of indirect PH signs in terms of the ventricles, pulmonary artery and right atrium. In the absence of TRV, mPAP was evaluated based on features learned from echocardiogram videos and Doppler images. In other words, we automate the process of PH probability assessment recommended in the guidelines~\cite{humbert20222022}. Notably, MePH facilitates a direct evaluation of mPAP to definitively classify PH rather than merely hinting at the possibility~\cite{ragnarsdottir2024deep,sun2024chamber,zuwei2024automatic,tison2019automated,bird2020assessment,dubrock2024electrocardiogram,liu2024deep,diller2022framework}. Our model demonstrates a significant narrowed limit of agreement compared with conventional method (see Figure~\ref{fig:diagnosis_result_mpap}a,b and Figure~\ref{fig:diagnosis_result_pvr}a,b), suggesting improved precision, which makes the application to an individual patient more accurate. These results further indicate MePH is a TRV-independent, objective and accurate diagnostic method.

Second, MePH is a useful adjunct in evaluating disease progression and treatment decisions. Of note, estimated pulmonary pressure is irrelevant to prognosis and therapeutic decision-making. The changes in pulmonary pressure do not necessarily reflect disease progression or improvement~\cite{humbert20222022}. In contrast, PVR is better than mPAP in estimating RV afterload and serves as an important hemodynamic parameter to assess disease progression and treatment response~\cite{humbert20222022,d2022echocardiographic,noordegraaf2019pathophysiology}. A number of Doppler echocardiography-derived non-invasive surrogates for PVR have been proposed, most of which were dependent on the combination of pulmonary artery pressure and flow assessment~\cite{naing2017non}. These parameters vary in accuracy and are not feasible in many conditions, such as severe tricuspid insufficiency, where the pressures in the RV and right atrium (RA) are almost equalized. Current studies showed improvements in right ventricle morphologic parameters such as RV end-diastolic area (RVEDA), left ventricular eccentricity index (LVEI), and RA area was in proportion to treatment-induced decrease in PVR. However, these improvements were only observed when there was a significant change in PVR~\cite{badagliacca2018prognostic,d2020risk}. Our model is capable of accurately assessing PVR, regardless of whether the resistance is mild-to-moderate or severely elevated. This requires the model to incorporate morphological features beyond the conventional measurements in the detection process. The activation heatmap in Figure~\ref{fig:cam_visual} further illustrates the rationale of the detection algorithm. The highlighted active regions focus more on the ventricular wall rather than the cardiac chamber dimension. A possible explanation is that subtle remodelling in the myocardium may occur before measurable chamber size changes. With enhanced sensitivity to changes in cardiac structure and function, MePH can help identify the pre-capillary component in the suspected PH cases and track the disease progression (i.e. change of PVR) during the entire diagnostic work-up.  

Third, our approach has the potential to optimize the classification and follow-up algorithm. By accurately assessing mPAP and PVR, MePH can help in the early detection of PH and evaluate the pre-capillary component in underserved communities. For suspected PH cases, PVR higher than 5 WU may indicate a severe pre-capillary component, which prompts the physician to refer the patient to PH centers for specialized care. For confirmed cases, MePH provides a non-invasive PVR assessment based on every 3-6 months of echocardiography follow-up. Once MePH detects significant disease progression, RHC may be indicated to confirm the results and adjust therapy. A satisfactory performance in tracking disease progression was demonstrated in an independent PH cohort over a 6-month observation, as shown in Figure~\ref{fig:diagnosis_result_pvr}c. Hemodynamic assessment is important in guiding treatment, while direct hemodynamic measurements cannot be performed in all patients or performed as routine examinations due to their invasive nature. Our model can prompt the timing of RHC and thus reduce the cost and the need for invasive procedures during the follow-up period. Our approach can also improve patient recruitment and minimize churn since it is based on routine echocardiography examinations without incurring any additional burden on patients.

Notably, the aim of this study was to develop an effective tool for the classification and follow-up of PH. We do not advocate for the replacement of RHC, the gold standard, nor to overshadow the crucial role of experienced experts in echocardiographic assessments. Our model serves as an adjunctive tool, providing a novel, non-invasive and accurate approach to assessing hemodynamic parameters. As shown in Figure~\ref{fig:overview}d, by integrating this model into the diagnostic workflow, we aim to enhance the precision and efficiency of PH management, enabling earlier interventions and more tailored treatment strategies. Furthermore, visualization of the deep learning model also deepens our understanding of disease. In this way, our study emphasizes the importance of innovative methodologies in enhancing our understanding and improving the management of this complex condition.

Limitations of this work are also important to acknowledge. Firstly, our paper focuses on PH measurements via echocardiography and cardiac hemodynamics. The training of MePH for multi-modal needs to train each modal independently and requires more time to converge. Due to types of ultrasound devices and scanning skill levels of echocardiographers being different across medical centers, the model is prone to overfit to training set, posing challenges in adapting our model to different scenarios or our-of-distribution groups. Secondly, we believe that other modalities that are able to reflect the cardiac function, such as Electrocardiogram~\cite{tison2019automated,bird2020assessment,dubrock2024electrocardiogram,liu2024deep} and Cardiac Auscultation~\cite{dennis2010noninvasive,kaddoura2016acoustic}. These modalities can be potentially utilised as an extra information for PH assessments. Our experiments have shown that incorporating more modalities can significantly improve performance. The introduction of different modalities can help the model understand cardiovascular and hemodynamics comprehensively. We will also further this study by incorporating advanced cardiac measurement techniques and more efficient modal fusion methods to acquire more accurate assessments.
For the application, our model requires the input echocardiography should only contain standard scanning views and conduct manually quality control to ensure the all data align with the experimental configuration. This is due to the training dataset only involve limited views of echocardiogram videos and Doppler images with a inclusion-exlusion standard as shown in supplementary Figure~\ref{fig:inclusion_exclusion_cascade}. In our future work, we will collect more data that cover more scanning angle of cardiac views. Moreover, we will built a quality control system to process input data automatically.

\fi 
\section*{Methods}
\label{sec:methods}
\subsection*{Dataset Description}
This study included suspected PH patients who underwent echocardiography at four hospitals in China (The Guangdong Cardiovascular Institute, Guangdong Provincial People’s Hospital, The First Affiliated Hospital of Guangzhou Medical University, Tongji University Affiliated Shanghai Pulmonary Hospital, and Fuwai Hospital Chinese Academy of Medical Sciences) between October 2020 and September 2023. The inclusion criteria were as follows: (1) patients suspected of PH who underwent transthoracic echocardiography; (2) patients scheduled for RHC; (3) the time interval between RHC and echocardiographic imaging did not exceed 24 hours; and (4) for patients with congenital heart disease (CHD) scheduled for transcatheter occlusion, echocardiographic imaging was performed within 24 hours before the procedure. The exclusion criteria were as follows: (1) patients with poor echocardiographic image quality or incomplete imaging data unsuitable for deep learning model construction; (2) patients for whom the interval between echocardiography and RHC exceeded 24 hours; and (3) patients who ultimately did not undergo RHC or had incomplete RHC data. A total of 1,000 suspected pulmonary hypertension patients who satisfied the requirement of the inclusion-exclusion cascade (see Supplementary Figure~\ref{fig:inclusion_exclusion_cascade}) were included in the cohort.

To validate the generalizability of our model, following the above inclusion and exclusion criteria, an independent external validation cohort of 237 patients was prospectively gathered from clinical trials conducted at eight other medical centers in China: Nanjing First Hospital, Tianjin Medical University General Hospital, The First Affiliated Hospital of Zhengzhou University, The First Affiliated Hospital of Xi'an Jiaotong University, Wuhan Asia Heart Hospital, Beijing Anzhen Hospital, Tongji Hospital Tongji Medical College of HUST and The General Hospital of the PLA Northern Theater Command. To verify the effectiveness of the proposed method in assessing changes in PVR and the progression of PH, we conducted a prospective study that included 110 patients who underwent RHC at initial diagnosis and a follow-up RHC three months later.

All patients underwent examinations utilizing commercially available echocardiography systems, specifically the EPIQ 7C by Philips, the Aloka 880 by HITACHI, the Vivid E9/E95 by GE in the internal dataset, and the EPIQ 7C by Philips and the Vivid E95 by GE in the external dataset. All systems are equipped with 3.5 MHz transducers, with imaging conducted in the left lateral decubitus position. In order to meet the inclusion-exclusion cascade illustrated in Supplementary Figure~\ref{fig:inclusion_exclusion_cascade}, echocardiography and RHC were required to be completed within a 24-hour window. Skilled operators performed all echocardiographic assessments in adherence to the guidelines set~\cite{humbert20222022}. 

Following the guildline~\cite{humbert20222022}, we selected echocardiogram videos from the PLAX, PSAX, and A4C views to help observe the echocardiographic signs related to PH, which are RV/LV basal diameter, flattening of the interventricular septum and TAPSE/SPAP ratio from the ventricles. Meanwhile, the selection of PALA from echocardiogram videos, RVOT, PR and PV views from Doppler images focuses on the diameter of the pulmonary artery and early diastolic pulmonary regurgitation velocity. In the final, the TR view from Doppler images provides the tricuspid regurgitation velocity for conventional mPAP and PVR calculation~\cite{chemla2004new,abbas2013noninvasive}. We ensure all echocardiogram videos have been taken over three cardiac cycles for clear observation. Patients with suboptimal imaging quality or those who did not undergo RHC within 24 hours were excluded from the study. Measurements included tricuspid and pulmonary valve regurgitation, with systolic pulmonary artery pressure (sPAP) and mean pulmonary artery pressure (mPAP) calculated using Bernoulli's equation. Additionally, right ventricular fractional area change (FAC), wall thickness (RVWT), tricuspid annular plane systolic excursion (TAPSE), and tricuspid annular peak systolic velocity (S') were assessed.

% \jw{Do we need to discuss the dataset? Collect from which center, how many cases, from which years to which years? Measurement of RHC, how echocardiographers scan the echocardiography, which technique they follow, and the types of devices.}
%
%
  \begin{figure}[t!]
    \centering
    \includegraphics[width=0.999\linewidth]{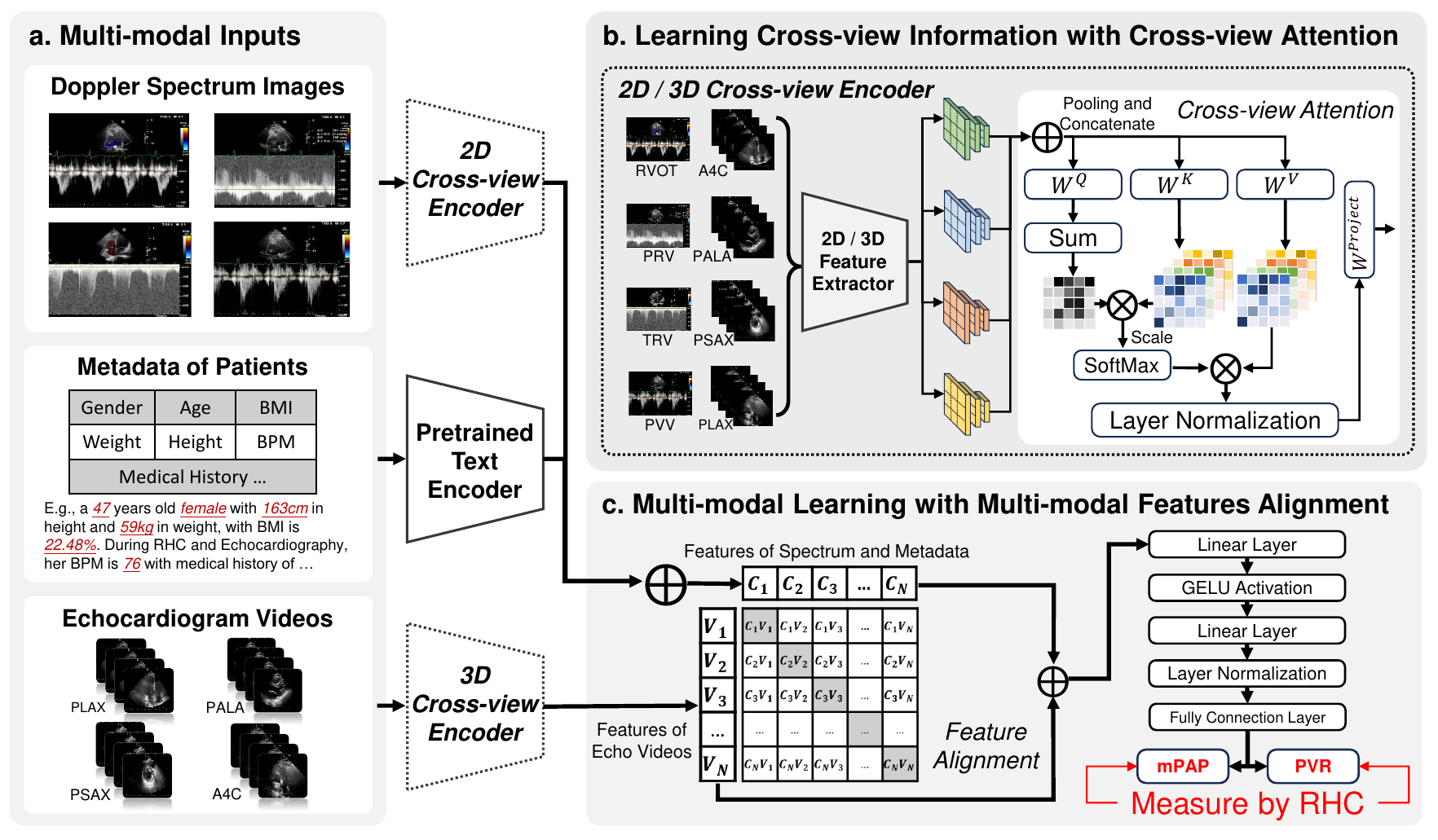}
    \caption{\textbf{Detail of our MePH framework.} \textbf{a, Multi-modal inputs} shows Doppler spectrum images (2D) and echocardiogram videos (3D) from different views are first fed into the cross-view encoder, utilizing 2D and 3D feature extractors, respectively. \textbf{b, Cross-view attention module} learns correlations and fuses features from each view for cardiac videos and spectral images, respectively.
    \textbf{c, Multi-modal feature alignment module} learns cross-modality feature alignment from 3D videos, 2D spectral images, and a text encoder derived from metadata. The combined features are then trained using an alignment loss and the MSE loss to facilitate the final assessment of mPAP and PVR, where the ground-truth mPAP and PVR are measured by RHC.
    }
    \label{fig:MePH_pipeline}
  \end{figure}

\subsection*{MePH Overview}

An overview of our MePH framework is depicted in Figure~\ref{fig:MePH_pipeline}. Initially, as shown in Figure~\ref{fig:MePH_pipeline}a, multi-view echocardiogram videos and Doppler images are processed through dedicated 3D and 2D cross-view encoders to extract relevant features. These encoders (illustrated in Figure~\ref{fig:MePH_pipeline}b) integrate information across various views, capturing detailed information on cardiac motion, structural characteristics, and hemodynamic parameters. Concurrently, a frozen pre-trained text encoder extracts patient metadata features essential for physical condition analysis. Following independent feature extraction from each modality, the system concatenates features derived from Doppler images with features obtained from patient metadata. The feature alignment module, as shown in Figure~\ref{fig:MePH_pipeline}c, aligns these concatenated features with those extracted from echocardiogram videos during the training stage. In our multi-modal data, echocardiogram videos provide comprehensive insights into cardiac motions and structures pertinent to PH. Meanwhile, Doppler images offer dynamic representations of hemodynamic conditions that reflect structural and motion abnormalities within the heart. Metadata further aids in distinguishing between PH patients under varying physiological states. By requiring each pair of Doppler images and corresponding metadata to match with appropriate echocardiogram videos during the alignment process, the model can construct a robust, more integrative representation for PH classification.
\\

\noindent \textbf{Learning Cross-view Information with Cross-view Attention.} In visual modalities, echocardiogram videos provide cardiac anatomical structures and motion, while Doppler images reflect cardiac hemodynamics. Furthermore, our PH dataset comprises multiple views for each case in both modalities, where these views provide a more comprehensive observation of cardiac behaviours. 
%Previous deep learning-based methods~\cite{ding2021mvfusfra,fantazzini20203d,liu2022transfusion,wang2022multi,zheng2023gl} usually fuse multi-view features by summation or concatenation. However, such strategies may lead to biased learning of multiple views. For example, the network towards learning features by a specific view presents a more direct cardiac behaviour from normal and PH cases. The learnt incomprehensive information may overfit the training dataset, which ignores potential relevant patterns from other views and leads to degraded performance. 
To fuse information across multiple views, we leverage cross-view attention to explore the consistent cardiac behaviour through shared information across views and filter out irrelevant features. Different from~\cite{ouyang2020video} that simply utilizes the spatiotemporal information for the regression task, as well as ~\cite{wang2024screening} concatenates features from different axes of CMR data, we propose a novel cross-view encoder that improves the accuracy of mPAP and PVR assessment. 

As shown in Figure~\ref{fig:MePH_pipeline}b, we utilize the same pipeline to learn cross-view information for both 2D and 3D input. Firstly, we consider cardiac motion from different scanned views to remain consistent within a case. Motion abnormalities across all cases can be attributed to several patterns. A parameter-shared Encoder is applied for feature extraction, which acts as a echocardiographer that examines cardiac motion across all views to explore abnormal cardiac behaviours. Moreover, a shared encoder significantly reduces the model parameters and avoids the gradient balancing requirement when optimizing among different Encoders. Hence, in the feature extraction stage, we employ the ResNet-3D~\cite{hara2018can} and ResNet-2D~\cite{he2016deep} backbones with 50 layers for video (3D) and image (2D) inputs, respectively. 3D and 2D inputs are encoded and pooled to the $\mathbb{R}^{C\times T\times W\times H}$ and $\mathbb{R}^{C\times W\times H}$ in shape, where $C$, $T$, $W$ and $H$ denote the channel, length, width and height. In our paper, shapes are set as $\mathbb{R}^{512\times 8\times 8\times 8}$ and $\mathbb{R}^{512\times 16\times 12}$ for 3D and 2D inputs.

As illustrated in Figure~\ref{fig:MePH_pipeline}b, the features from all input views are concatenated and projected to Query ($\mathbf{Q}$), Key ($\mathbf{K}$) and Value ($\mathbf{V}$) vectors by independent linear transformation $\mathbf{W^{Q}}$, $\mathbf{W^{K}}$ and $\mathbf{W^{V}}$. Then, we sum the $\mathbf{Q}$ across views to compress and merge features. The key $\mathbf{K}$ responds to the query $\mathbf{Q}$ to search the related information associated with candidate features in $\mathbf{V}$. Since both $\mathbf{K}$ and $\mathbf{V}$ are not compressed, $\mathbf{Q}$ can map and pay attention to related cardiac behaviours across views both temporally and spatially. The attentive output $\mathbf{F_A}$ is formulated as follows:
\begin{equation}
  \mathbf{F_A}=\text{SoftMax}\left(\frac{\mathbf{(\sum_{i=1}^{4}Q_i)}\mathbf{K}^{\mathsf{T}}}{\sqrt{d_k}}\right)\mathbf{V},
\end{equation}
where $d_k=C$ is the scaling factor to control the variance of the dot product. $\mathbf{F_A}$ then passes to the normalization layer and is projected with a linear layer with weight $\mathbf{W^\text{project}}$. In this way, the model can learn similar cardiac behaviours across different views without manually weighing their correlations. Moreover, we randomly mask views during training, which allows the model to assess the result with inadequate views. The supplementary Figure~\ref{fig:ablation}b shows the ablation study of PH assessment with a single view as input. The result illustrates that different views do not equally contribute to the model, as they may present different cardiac motion and structural information that is related to the PH. This module helps MePH accurately capture related patterns and extract features under diverse input conditions, improving its robustness and generalizability.
\\

\noindent \textbf{Multi-modal Learning with Multi-modal Features Alignment.} Clinically, echocardiographers usually utilize echocardiogram videos for PH assessment, while the metadata and Doppler images serve as assistant modalities for reference. Moreover, as shown in Supplementary Figure~\ref{fig:ablation}, using only video modality can reach an AUC of 0.842 in PH assessment via mPAP. However, when PH is assessed by only Doppler images, this number drops to 0.715. These results indicate that solely using a single modality is not able to reflect the actual characteristics of PH. Furthermore, Doppler images describe cardiac information from the view of hemodynamics, which is seen as a language of cardiac motion. For example, tricuspid regurgitation occurs when the valve between the right ventricle and right atrium does not close properly, which is a common cardiac abnormality in patients with PH. This abnormality can be observed in echocardiogram videos, and the Doppler image can detect the intensity of tricuspid regurgitation. Likewise, metadata often reflects the physical statistics of the patient demographics, which represent possible physiological information about patients with PH and cannot accurately reflect their cardiac function. For PH classification and assessment, the Doppler images and metadata are the profiles of the echocardiogram videos. Compared to directly conducting summation or concatenating the multi-modal information, we consider that alignment is a more appropriate way to learn the correlation across modalities. 

In comparison to~\cite{christensen2024vision}, which aligns the text and vision feature for an echocardiography vision-language model. We apply a different strategy for modalities alignment as shown in Figure~\ref{fig:MePH_pipeline}c. When given a batch of N (video, image, text) pairs, we start by combining image and text features to create \texttt{(video, [image, text])} pairs (text features are extracted by pretrained DistilBERT~\cite{sanh2019distilbert}). In this process, the echocardiogram videos and concatenated features are represented as $\mathbf{V}_j$ and $\mathbf{C}_j$, where $j$ refers to the $j$-th sample in a batch. Subsequently, our model is trained to identify which of the $N \times N$ potential \texttt{(video, [image, text])} pairings within a batch actually occurred. To achieve this, our model learns a multi-modal embedding space through the joint training of 2D and 3D cross-view encoders, along with a fixed pre-trained text encoder. The goal is to maximize the cosine similarity of the \texttt{video} and \texttt{[image, text]} embeddings for the $N$ real pairs in the batch, while simultaneously minimizing the cosine similarity of the embeddings for the $N^2 - N$ incorrect pairings. For the optimization process, we utilize a symmetric cross-entropy loss over these similarity scores, which is formulated as:
\begin{equation}
    \mathcal{L}(\mathbf{V},\mathbf{C}) =-\sum\nolimits^{N,N}_{j\le N,k\le N}\log \left(\frac{\exp \left(\mathbf{V}_j{\mathbf{C}_k}^{\mathsf{T}} / \tau\right)}{\sum_{j \neq k} \exp \left(\mathbf{V}_j{\mathbf{C}_k}^{\mathsf{T}} / \tau\right)+\exp \left(\mathbf{V}_j{\mathbf{C}_k}^{\mathsf{T}} / \tau\right)}\right),
\label{eq:alignment}
\end{equation}
where temperature $\tau$ controls the strength of penalties on hard negative pairs, $j$ and $k$ indicate the $j$-th video feature and $k$-th concatenated feature of the $N\times N$ possible pairings.

We further concatenate $\mathbf{V}$ and $\mathbf{C}$, as the outputs of the three modalities are expected to be both mPAP and PVR assessment basis. Then, a multilayer perception, as shown in Figure~\ref{fig:MePH_pipeline}c, is designed to assess the mPAP and PVR. For the optimization, we apply supervision on both the final result and the branch outputs. In this paper, the mean squared error loss is applied to measure the distance between the assessment mPAP, PVR and the ground-truth mPAP$_\text{RHC}$, PVR$_\text{RHC}$. As shown in Table~\ref{table:dataset_statics}, the mPAP usually has a higher value compared with the PVR. Directly adding their loss leads to the mPAP dominating the training. Hence, a hyper-parameter $\lambda$ that is set as $0.25$ is introduced to balance them in a batch as the following formulation:
\begin{equation}
    \mathcal{L}_\text{MSE} = \frac{1}{N}\left(\lambda\sum^N(\text{mPAP} - \text{mPAP}_\text{RHC})^2 + \sum^N(\text{PVR} - \text{PVR}_\text{RHC})^2\right).
\end{equation}
The overall loss of MePH training is then formulated as $\mathcal{L}_\text{total}=0.1\mathcal{L}(\mathbf{V},\mathbf{C})+\mathcal{L}_\text{MSE}$.
\\

\noindent \textbf{Training Details.} For both 2D and 3D cross-view encoders, we use the Adam optimizer with default hyperparameter settings. Then, we set an initial learning rate of $3\times10^{-4}$, applying the cosine decayed to zero over the course of the training run. We used a batch size of 16 and trained on four Nvidia RTX3090 24GB GPUs for approximately 5 days with a total of 300 epochs. In total, 900 cases are randomly divided into 800 and 100 for training and validation. The trained model in the epoch with the lowest MAE of mPAP and PVR on the validation set was saved and selected for both internal and external testing. For the feature alignment during training, we divide features by their norms to ensure they stay in the same magnitude. The cosine similarity metric computed between features of two modalities always returns a value in the range of $-1$ to $1$.

% \subsubsection*{Observer study}
%
% \jw{The clinical applicability of the model was evaluated in an observer study with ? echocardiographers on the external data. ? junior echocardiography echocardiographers (name?., with ? years of experience, respectively) and ? senior echocardiography echocardiographers (name?., with ? years of experience, respectively) who were blind to the reference standard independently performed Echocardiography interpretations. They viewed both Echocardiogram videos and Doppler Doppler images using 3D Slicer platform~\cite{fedorov20123d} (Version 5.6.2, 2024-04-05) and performed classifications for symptoms of PH. After a washout period of ? days, they performed repeated interpretations with model assistance. The comparison was performed between echocardiography echocardiographers with and without model assistance.}
%

\if 1 
\subsection*{Previous Works on Automatic Pulmonary Hypertension Assessment and Classification}
Echocardiography, as the most commonly used diagnostic tool, can provide high-resolution imaging for cardiac anatomical structures and motion. Studies on echocardiography utilising deep learning have been explored in analysing cardiac structures and their function~\cite{ouyang2020video,ghorbani2020deep,duffy2022high,he2023blinded,holste2023severe,yuan2023deep,christensen2024vision}. Recent deep learning-based research related to PH mainly focuses on classifying PH and non-PH groups by using Echocardiography~\cite{ragnarsdottir2024deep,sun2024chamber,zuwei2024automatic}. Ragnarsdottir, \emph{et al.}~\cite{ragnarsdottir2024deep} collect a dataset with multi-view echocardiogram videos and specifically design a multi-view fusion network for newborn PH classification. %They manually classify the newborns with PH by calculating the echocardiography characteristics instead of RHC measurement. 
Sun, \emph{et al.}~\cite{sun2024chamber} propose an attention network for PH classification through learning chamber information from single-view 2D echocardiogram images. 
Diller, \emph{et al.}~\cite{diller2022framework} propose a network that fuses echocardiogram video features from two views. The above approaches train on PH data classification according to echocardiography characteristics solely with the absence of RHC measurement.
%They conduct experiments with data collected from adult PH patients diagnosed through Doppler echocardiography. 
Liao, \emph{et al.}~\cite{zuwei2024automatic} propose a U-Net-based network to classify PH patients by RHC from single-view echocardiogram videos. 

%Similar to the above methods, research~\cite{tison2019automated,bird2020assessment,dubrock2024electrocardiogram} designed network only focus on ECG, while Liu, \emph{et al.}~\cite{liu2024deep} jointly utilize ECG and chest X-Ray for PH diagnosis by specify design a network that fuses two modalities. 
%

The above works mainly focused on a single modality without considering the utilization of Doppler images and patient metadata. Also, the lack of learning patterns from multiple scanned views and the temporal sequence present by echocardiogram videos may impede the comprehensive learning of cardiac structures and motion. Though Liu, \emph{et al.}~\cite{liu2024deep} have utilized multi-modal biomedical data, ECG and Chest X-ray presented only limited cardiac information, resulting in unsatisfactory performance. Existing research explores their method with data from patients classified by echocardiography characteristics. However, such an approach may produce a huge bias compared to RHC measurement (see Results in Figures~\ref{fig:diagnosis_result_mpap} and~\ref{fig:diagnosis_result_pvr}), indicating the model has the potential to misclassify PH patients when deployed in real clinical scenarios. In the PH classification task, these studies solely consider classifying symptoms without the central element mPAP and PVR of PH, offering limited assistance for subsequent clinical treatment. Our work is built on the multi-modal data that can learn the more comprehensive pattern of PH. The RHC annotation not only allows the model to classify the PH but also provides a direct measurement of mPAP and PVR.

\fi 

\subsection*{Data Pre-processing and Augmentation}

Echocardiography encompasses two modalities: multi-view echocardiogram videos and multi-view Doppler spectrum images.
Frames from the \textbf{echocardiogram videos} are first cropped to retain only the cardiac beats and remove unnecessary watermarks. We then equidistantly sample the video to 128 frames, resize them, and perform random cropping to dimensions of $256 \times 256$ pixels in height and width. For videos with insufficient frames, we fill in empty frames to achieve the target length. During training, we set a random offset ratio of up to 30\% for cropping frames. Additionally, we randomly apply horizontal flipping for temporal augmentation and rotation for spatial variation.
Since devices from different manufacturers have varying imaging parameters, such as color, brightness, and contrast, these discrepancies can lead to feature shifts and degrade model performance when applied to unseen data from other medical centers. To address this issue, we apply random color jitter to introduce variability in low-level information and normalize the values to the range of [0, 1]. In the inference stage, we resize and apply fixed cropping to the input without any random transformations.
\textbf{Doppler images} include both cardiac scans and hemodynamic information. Non-proportional transformations can alter the information in the Doppler images, so we only perform proportional scaling, ensuring that all inputs with RGB channels are maintained at dimensions of $800\times 600$ pixels in width and height. Similar to the augmentation applied to echocardiogram videos, color jitter and normalization are randomly applied during training.
\textbf{Metadata of patients}, as shown in Figure~\ref{fig:MePH_pipeline}a, consists only of text content related to patients' physical conditions, such as weight, gender, age, etc. To extract features from the metadata, we use a Byte-Pair Encoding tokenizer pre-trained on the GPT-2 data corpus, which tokenizes all text and encodes the metadata with an average of 107.3 (±25.1) tokens per case.

The parameter-based learning models, including Multi-Layer Perceptron (MLP), BERT~\cite{devlin2019bert}, DistilBERT~\cite{sanh2019distilbert}, utilize only structured echocardiographic parameters as input features. For the MLP model, we employed two fully connected layers with 256 hidden units. For the BERT-based model, we fine-tuned the BERT base model (uncased) pretrained on English corpora using a masked language modelling (MLM) objective. Both the MLP and BERT models were adapted to accept echocardiographic parameters as input and trained to predict ground-truth mPAP and PVR values measured via RHC. As detailed in Supplementary Table~\ref{table:dataset_statics}, a total of seven echocardiographic features (RVW, TAPSE, S', FAC, Echo-RAP, VTI, and TRV) and two EHR features (Sex, Age) were used. To ensure consistency and fair comparison with the proposed vision-language model, we adopted identical training and validation datasets as MePH, selected the model with the best validation performance, and evaluated its generalizability on both internal and external test cohorts.

\subsection*{Clinical Taxonomy}
Patients with PH are clinically categorized into three severity levels based on mean pulmonary arterial pressure (mPAP) measured in mmHg, as outlined in the guideline~\cite{humbert20222022}. According to this standard, PH is classified as severe ($\text{mPAP} \ge 50\;\text{mmHg}$), moderate ($50\;\text{mmHg} > \text{mPAP} \ge 35\;\text{mmHg}$), and mild ($35\;\text{mmHg} > \text{mPAP} \ge 20\;\text{mmHg}$). Additionally, the guideline provides criteria for assessing PH severity using pulmonary vascular resistance (PVR), defining severe cases as those with $\text{PVR} \ge 5\;\text{WU}$ and moderate-mild cases as having $5\;\text{WU} > \text{PVR} \ge 2\;\text{WU}$. To differentiate between PH patients and healthy individuals, we also gather data from normal subjects ($\text{mPAP} < 20\;\text{mmHg}$ and $\text{PVR} < 2\;\text{WU}$).

echocardiographers provided manual mPAP$_\text{Echo}$ and PVR$_\text{Echo}$ assessments for PH obtained by echocardiography. According to the research and guidelines~\cite{chemla2004new,humbert20222022}, the mPAP$_\text{Echo}$ is calculated from maximum tricuspid regurgitation velocity (TRV$_\text{max}$) and estimated right atrial pressure (eRAP) with the following formulation:
\begin{equation}
    \text{mPAP}_\text{Echo} = 0.61 \times (4\times\text{TRV}^2_\text{max} + \text{eRAP}) + 2.
\end{equation}
For the calculation of PVR, refer to the research~\cite{abbas2013noninvasive}, we utilize the Tricuspid Regurgitation Velocity (TRV) and the time-velocity integral of the right ventricular outflow tract (TVI$_\text{RVOT}$), which is formulated as:
\begin{equation}
    \text{PVR}_\text{Echo} = 5.19 \times \frac{\text{TRV}^2}{\text{TVI}_\text{RVOT}} - 0.4.
\end{equation}
The units of the $\text{mPAP}_\text{Echo}$ and $\text{PVR}_\text{Echo}$ are the manometric unit of pressure (mmHg) and the Wood (WU), respectively.
\subsection*{Ethical and Information Governance Approvals}
This research was approved by the Institutional Review Board of Guangdong Cardiovascular Institute, Guangdong Provincial People’s Hospital (\textit{Approval No.} KY2023-770-02). This research and the relevant data were also proved by all cooperating medical centers. Each case and its metadata were anonymized, and all personally identifiable information was removed to prevent any leakage of identifying details. To ensure the dataset is error-free and complete and to confirm the absence of any personally identifiable information, all echocardiogram videos and Doppler images were double-checked by three echocardiographers with over 8 years of experience. This process ensures data integrity and patient confidentiality.

\subsubsection*{Statistical analysis}
In this paper, the Kruskal-Wallis test was used to compare the ages of patients across different centers, while the chi-square test was applied to compare the distribution of sexes. The Bland-Altman plots referenced as Figures~\ref{fig:diagnosis_result_mpap} and~\ref{fig:diagnosis_result_pvr} are employed for assessing the concordance between AI assessments and RHC measurements in both internal and external testing sets. We calculate the $95\%$ limits of agreement for each comparison, which involve the average difference ± 1.96 standard deviations of the difference. If not otherwise stated, the significance of the
difference in AUC-ROC and MAE is validated by Delong-test~\cite{delong1988comparing} and T-test, respectively, with $P\leq0.05$ as significant.

\section*{Code Availability}
All codes were conducted in Python using the PyTorch deep-learning library used in this study. All codes and models for reproducing this study will be publicly available after the paper is accepted. 
%Please see our code and model here: \href{https://anonymous.4open.science/r/MePH/}
%{https://anonymous.4open.science/r/MePH/}. 
Supplementary Figure~\ref{fig:demo} illustrates the clinical application interface of our system, integrating echocardiogram video, Doppler images and EHR data with automatic PAH assessment and AI-guided decision support. 
%Please visit our Demo~\href{https://drive.google.com/file/d/1tx4RBkKjT5uxbiqZp7N7fzJN\_LmPffZq/view}{https://drive.google.com/file/d/1tx4RBkKjT5uxbiqZp7N7fzJN\_LmPffZq/view} for more information on how echocardiographers use the platform.
% All code for reproducing this study can be found at  \xmli{ xxxxxx.}

\section*{Data Availability}
The raw data collected and processed in this study are supervised under corresponding institutions. The data are available under restricted access, which can be obtained by emailing the corresponding author with all requests for academic use. The requirements concerning institutional policies will be evaluated, and data can only be shared for non-commercial academic usage with a formal material transfer agreement. All requests will be promptly reviewed within a timeframe of 15 working days. The data generated in this study is provided in the Source Data file. The data that helps to reproduce this study are available in the code repository. Source data are provided in this paper.
% \xmli{Our internal dataset, including training and testing set, could be released.}

\medskip

{
\small
\bibliographystyle{unsrt} 
\bibliography{main}
}

%%%%%%%%%%%%%%%%%%%%%%%%%%%%%%%%%%%%%%%%%%%%%%%%%%%%%%%%%%%%
\clearpage
\newpage

\appendix

\section*{Acknowledgements}

This work was supported by grants from the National Natural Science Foundation of China (Grant No. 62306254), the Research Grants Council of the Hong Kong Special Administrative Region (Project Reference Number: T45-401/22-N). 
The authors are very grateful to the patients who contributed to the echocardiography and RHC data in this paper. This study was also supported by the National Natural Science Foundation of China (82371963), Guangdong Basic and Applied Basic Research Foundation (2023A1515011366, 2023A1515111176), the NSFC launching fund of the Guangdong Provincial People’s Hospital (8227070211), Guangdong Medical Science and Technology Research Foundation (B2023010), and Guangzhou Basic and Applied Basic Research Special Youth Doctoral "Setting Sail" Project (2024A04J5038).

\section*{Author Contributions Statement}

X.M.-L. and H.W.-F. conceived and designed the study. J.W.-Y. conducted the experiments. J.W.-Y., B.-P., and X.M.-L. form the technical group, which oversees the research framework, algorithm design, experimentation, and application development. Specifically, J.W.-Y. and X.M.-L. led the research efforts, algorithm design, and experiments, ensuring the reproducibility of all results. B.-P. contributed to algorithm design and experiments by providing valuable suggestions and feedback for the manuscript. J.R.-G helped design the interface and the demo of the PH diagnosis platform.
The medical group includes T.R.-H., X.W.-X., S.W.-D., Q.H.-Z., Y.-J., J.X.-Z., C.J.-Z., and H.W.-F. This group is responsible for data collection, annotation, clinical protocol formulation, and manual evaluation. T.R.-H., S.W.-D., Q.H.-Z., and Y.-J. focused on scanning echocardiogram videos and Doppler spectrum images. T.R.-H., X.W.-X., and J.X.-Z. measured echocardiography parameters and organized the collected data. C.J.-Z. and H.W.-F. performed right heart catheterization (RHC) on patients and contributed to the development of clinical protocols. Additionally, T.R.-H. and H.W.-F. played a key role in designing the experiments.
J.W.-Y., T.R.-H., S.W.-D., and X.W.-X. contributed equally to this work.

% Draft writing is mainly from J.W.-Y., T.R.-H., H.W.-F. and X.M.-L. 

\section*{Competing Interests}
The authors declare no competing interests.
\section*{Additional Information}
%

% \subsection*{Supplementary information}
%

%
\subsection*{Correspondence}
X.M.-L. and H.W.-F. are the corresponding authors: eexmli@ust.hk, feihongwen@gdph.org.cn.

% \subsection*{Requests for Materials}

\newpage
\begin{supptable}[t!]
  % \vspace{10pt}
  \centering
  \caption{Clinical characteristics across medical center study cohorts, reported per echocardiography study.}
  \begin{adjustbox}{width=0.999\textwidth}
  \begin{threeparttable}
  \setlength{\tabcolsep}{30pt}
  \begin{tabular}{l c c}
\hline
 & \multicolumn{1}{c}{\textbf{Internal Dataset (n=1,000)}} & \multicolumn{1}{c}{\textbf{External Dataset (n=237)}} \\ 
\hline \hline
\textbf{Demographics} & & \\ 
Sex (Female \%) & 622 (62.20\%) & 197 (83.1\%) \\  
Age (year) & 43 $\pm$ 15 (14 - 77) & 40.747 $\pm$ 11.653 (19 - 67) \\ 
\hline
\textbf{Echocardiography} & Mean $\pm$ Std (Range) & Mean $\pm$ Std (Range) \\ 
RVW (mm) & 5.596 $\pm$ 2.259 (2.6 - 19.0) & 5.928 $\pm$ 1.721 (3.0 - 11.0) \\ 
TAPSE (mm) & 19.37 $\pm$ 5.825 (8.0 - 45.0) & 15.896 $\pm$ 4.298 (2.0 - 29.0) \\ 
S' (cm/s) & 11.84 $\pm$ 3.115 (5.40 - 22.0) & 10.826 $\pm$ 2.570 (3.60 - 17.9)\\ 
FAC (\%)  & 35.66 $\pm$ 9.033 (6.0 - 57.0) & 28.961 $\pm$ 11.246 (7.10 - 57.6)\\ 
TRV (m/s)  & 3.682 $\pm$ 0.944 (1.0 - 7.25) & 4.175 $\pm$ 0.873 (1.90 - 6.75)\\ 
VTI (cm)  & 13.53 $\pm$ 4.952 (2.90 - 36.5) & 12.24 $\pm$ 4.284 (4.10 - 26.4)\\ 
Echo-PVR (WU)  & 5.874 $\pm$ 4.970 (0.45 - 33.2) & 8.506 $\pm$ 4.964 (0.70 - 47.04) \\ 
Echo-mPAP (mmHg)  & 41.43 $\pm$ 18.60 (2.0 - 135.1) & 50.68 $\pm$ 18.456 (12.64  - 118.05) \\ 
\hline
\textbf{Right Heart Catheter} & Mean $\pm$ Std (Range) & Mean $\pm$ Std (Range) \\ 
mRAP (mmHg)  & 6.184 $\pm$ 3.98 (0.0 - 32.0) & 8.750 $\pm$ 5.134 (1.0 - 28.0)\\ 
sPAP (mmHg)  & 66.70 $\pm$ 29.99 (10.0 - 168.0) & 80.236 $\pm$ 24.994 (29.0 - 172.0) \\ 
mPAP (mmHg)  & 37.40 $\pm$ 17.79 (5.0 - 107.0) & 50.785 $\pm$ 16.356 (20.0 - 127.0) \\ 
PVR (WU)   & 7.887 $\pm$ 5.940 (0.66 - 41.54) & 9.581 $\pm$ 5.027 (1.35 - 26.67) \\ 
PAWP (mmHg)  & 8.102 $\pm$ 4.441 (0.0 - 29.0) & 10.737 $\pm$ 3.945 (2.0 - 30.0) \\ 
\hline
\textbf{Device Type} & Number (Ratio) & Number (Ratio) \\
PHILIPS & 727 (72.7\%) & 73 (30.80\%) \\
GE & 155 (15.5\%) & 164 (69.20\%) \\
ALOKA & 118 (11.8\%) & 0 (0\%) \\
\hline
\end{tabular}
  \begin{tablenotes}[flushleft]\footnotesize
    \item[*]Abbreviations: Echo, echocardiography estimated value; FAC, fractional area change; TRV, tricuspid regurgitation velocity; VTI, velocity time integral; mPAP, mean pulmonary arterial pressure; mRAP, mean right atrium pressure; PAWP, pulmonary arterial wedge pressure; PH, pulmonary hypertension; PVR, pulmonary vascular resistance; RAP, mean right atrial pressure; RVWT, right ventricular wall; S’, tricuspid valve annulus peak systolic velocity; sPAP, systolic pulmonary arterial pressure; TAPSE, tricuspid annular plane systolic excursion.
  \end{tablenotes}
  \end{threeparttable}
  \end{adjustbox}
\label{table:dataset_statics}
\end{supptable}
\clearpage

\newpage
\begin{supptable}[t]
\centering
\caption{\textbf{The Etiology of PH from internal and external datasets}. PH can be classified into various subtypes based on their pathological, pathophysiological, and therapeutic characteristics. In our study, we analyzed both internal and external datasets to quantify the prevalence of different types of PH. Healthy controls are defined as individuals who do not have the disorder or disease associated with the specific type of PH under ultrasound investigation.}
\begin{adjustbox}{width=0.999\textwidth}
\begin{tabular}{l c c}
\hline
\textbf{PH Type} & \textbf{Abbreviation} & \textbf{Number} \\ \hline
\multicolumn{3}{c}{\textbf{Internal Dataset}}\\ \hline \hline
Healthy Controls & None-PH & 170 \\ 
Idiopathic Pulmonary Arterial Hypertension & IPAH & 199 \\   
Pulmonary Arterial Hypertension Associated with Connective Tissue Disease & CTD-PAH & 8 \\ 
Pulmonary Arterial Hypertension Associated with Congenital Heart Disease & CHD-PAH & 211 \\ 
Pulmonary Hypertension Associated with Left Heart Disease & LHD & 27 \\
Chronic Thrombo-embolic Pulmonary Hypertension & CTEPH & 385 \\
\hline
\multicolumn{3}{c}{\textbf{External Dataset}}\\ \hline \hline
Idiopathic Pulmonary Arterial Hypertension & IPAH & 127 \\  
Hereditary Pulmonary Arterial Hypertension & HPAH & 4 \\  
Pulmonary Arterial Hypertension Associated with Connective Tissue Disease & CTD-PAH & 57 \\ 
Pulmonary Arterial Hypertension Associated with Congenital Heart Disease & CHD-PAH & 43 \\ 
Pulmonary Arterial Hypertension Associated with Portal Hypertension & PoPH & 6 \\ 
\hline
\end{tabular}
\end{adjustbox}
\label{tab:ph_etiology}
\end{supptable}
\clearpage

\newpage
\begin{supptable}[t!]
\centering
\caption{Performance comparison of Vision-language based model MePH, Echocardiography parameter-based baseline models and Echocardiograher on mPAP and PVR prediction. For the assessment of mPAP and PVR, we have reported mean absolute error (MAE), coefficient of determination ($R^2$), and root mean squared error (RMSE) in both internal and external datasets. Meanwhile, the GIGA floating-point operations per second (GFLOPS), computational time (milliseconds), and total parameters indicate the model's inference efficiency.}
\begin{adjustbox}{width=0.999\textwidth}
\begin{tabular}{lccccccc}
\toprule
\textbf{Task} & \textbf{Method} & \textbf{MAE ↓} & $\mathbf{R^2}$ ↑ & \textbf{RMSE ↓} & \textbf{GFLOPS} & \textbf{TIME (ms)} & \textbf{Params (M)} \\
\midrule

\multicolumn{8}{c}{\textbf{Internal Dataset}} \\
\midrule \midrule
\multirow{4}{*}{mPAP} 
    & Echocardiograher & 9.609 & 0.305 & 13.47 & - & - & - \\
    & MLP         & 7.561 & 0.598 & 10.25 & $6.144e-06$ & 1.232 & 0.064 \\
    & BERT-base   & 6.981 & 0.644 & 9.650 & 12.58 & 34.99 & 102.5 \\
    & BERT-large  & 7.423 & 0.621 & 9.954 & 47.75 & 74.12 & 335.4 \\\
    & DistilBERT  & 7.097 & 0.635 & 9.771 & 5.952 & 24.70 & 66.61 \\
    & \textbf{MePH}        & \textbf{5.099} & \textbf{0.832} & \textbf{7.751} & 904.98 & 45.21 & 194.3  \\
\midrule
\multirow{4}{*}{PVR} 
    & Echocardiograher & 2.769 & 0.296 & 4.196 & - & - & - \\
    & MLP         & 2.399 & 0.504 & 3.606 & $6.144e-06$ & 1.232 & 0.064 \\
    & BERT-base   & 2.486 & 0.443 & 3.822 & 12.58 & 34.99 & 102.5 \\
    & BERT-large  & 2.525 & 0.490 & 3.658 & 47.75 & 74.12 & 335.4 \\
    & DistilBERT  & 2.531 & 0.453 & 3.787 & 5.952 & 24.70 & 66.61 \\
    & \textbf{MePH}        & \textbf{1.571} & \textbf{0.691} & \textbf{2.566} & 904.98 & 45.21 & 194.3  \\

\midrule
\multicolumn{8}{c}{\textbf{External Dataset}} \\
\midrule \midrule
\multirow{4}{*}{mPAP} 
    & Echocardiograher & 10.30 & 0.259 & 14.05 & - & - & - \\
    & MLP         & 13.02 & 0.040 & 17.11 & $6.144e-06$ & 1.232 & 0.064 \\
    & BERT-base   & 11.74 & 0.131 & 16.27 & 12.58 & 34.99 & 102.5 \\
    & BERT-large  & 12.73 & 0.039 & 17.16 & 47.75 & 74.12 & 335.4 \\
    & DistilBERT  & 12.58 & 0.052 & 16.99 & 5.952 & 24.70 & 66.61 \\
    & \textbf{MePH}        & \textbf{9.162} & \textbf{0.619} & \textbf{12.45} & 904.98 & 45.21 & 194.3  \\
\midrule
\multirow{4}{*}{PVR} 
    & Echocardiograher & 4.224 & -0.099 & 5.380 & - & - & - \\
    & MLP         & 3.253 & 0.305 & 4.357 & $6.144e-06$ & 1.232 & 0.064 \\
    & BERT-base   & 3.420 & 0.222 & 4.608 & 12.58 & 34.99 & 102.5 \\
    & BERT-large  & 3.383 & 0.248 & 4.531 & 47.75 & 74.12 & 335.4 \\
    & DistilBERT  & 3.421 & 0.230 & 4.584 & 5.952 & 24.70 & 66.61 \\
    & \textbf{MePH}        & \textbf{3.147} & \textbf{0.582} & \textbf{3.640} & 904.98 & 45.21 & 194.3 \\

\bottomrule
\end{tabular}
\end{adjustbox}
\label{tab:performance_across_four_methods}
\end{supptable}
\clearpage

\newpage
\begin{suppfigure}[t]
\centering
\includegraphics[width=\textwidth]{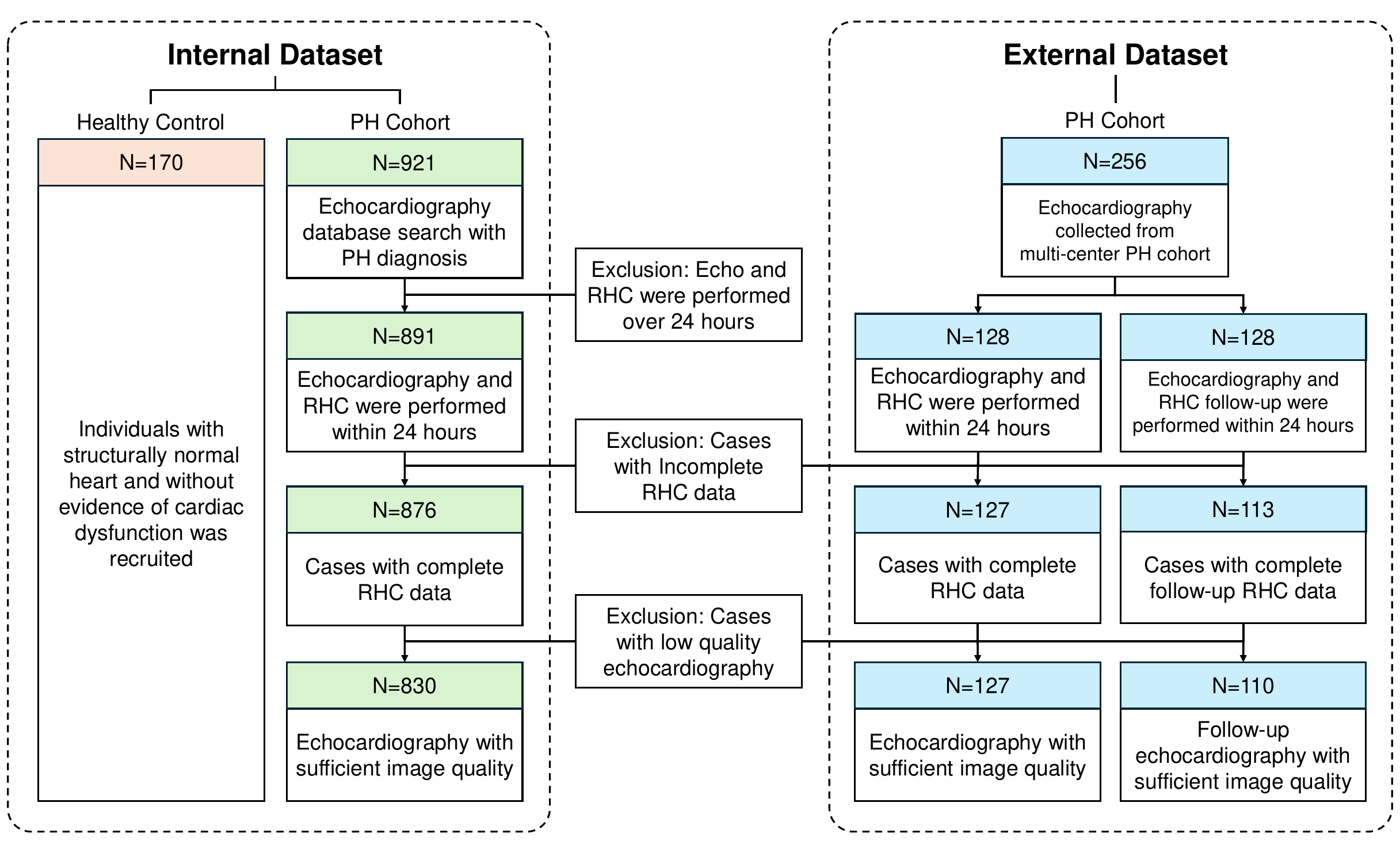}
\caption{\textbf{Inclusion-exclusion cascade for the PH and the healthy control cohorts.}}
\label{fig:inclusion_exclusion_cascade}
\end{suppfigure}
\clearpage

\newpage
\begin{suppfigure}[t]
\centering
\includegraphics[width=0.999\textwidth]{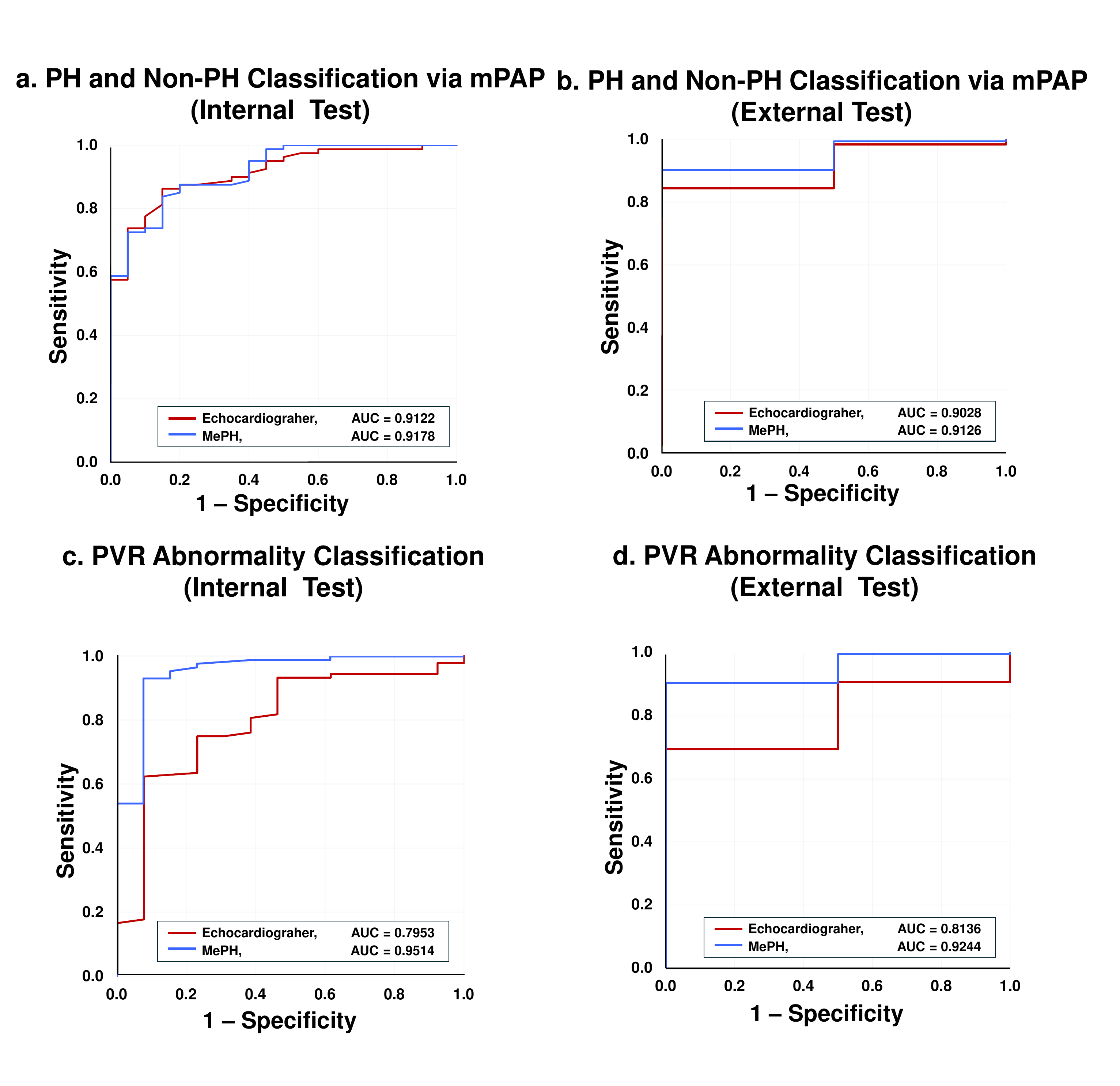}
\caption{\textbf{a,} The AUC-ROC of PH and Non-PH results are classified by mPAP (mPAP$<20$ as Non-PH and PVR$\ge2$ as PH) from the internal test set. \textbf{b,} The AUC-ROC of PH and Non-PH results are classified by mPAP (mPAP$<20$ as Non-PH and PVR$\ge2$ as PH) from the external test set. \textbf{c,} The AUC-ROC of PVR abnormality (PVR$<2$ as normal and PVR$\ge2$ as abnormal) is classified by PVR from the internal test set. \textbf{d,} The AUC-ROC of PVR abnormality (PVR$<2$ as normal and PVR$\ge2$ as abnormal) results is classified by PVR from the external test set.}
\label{fig:classification_result}
\end{suppfigure}
\clearpage

% \newpage
% %
% \begin{suppfigure}[t]
% \centering
% \includegraphics[width=\textwidth]{SuppFigure/Disease_types.pdf}
% \caption{\textbf{Performance comparison across different subtypes of Pulmonary Hypertension}. \textbf{a,} The pie chart shows the overall performance across different PH subtypes. All results in tables of each PH subtype were reported by mean absolute error (MAE) and the coefficient of determination $R^2$ for PH classification based on mPAP (mmHg) and PH progression assessment based on PVR (WU). \textbf{b,} The performance of mPAP assessment in the IPAH (n=127), CTD-PAH (n=57), CHD-PAH (n=43) and others (HPAH (n=4) and PoPH (n=6)). \textbf{c,} The performance of PVR assessment in the IPAH (n=127), CTD-PAH (n=57), CHD-PAH (n=43) and others (HPAH (n=4) and PoPH (n=6)). All results were reported in the external dataset. For the full names of these Pulmonary Hypertension subtypes, please see Supplementary Table~\ref{tab:ph_etiology}.}
% \label{fig:cross_disease_subtype_result}
% \end{suppfigure}
% %
% \clearpage

\newpage
\begin{suppfigure}[t]
\centering
\includegraphics[width=\textwidth]{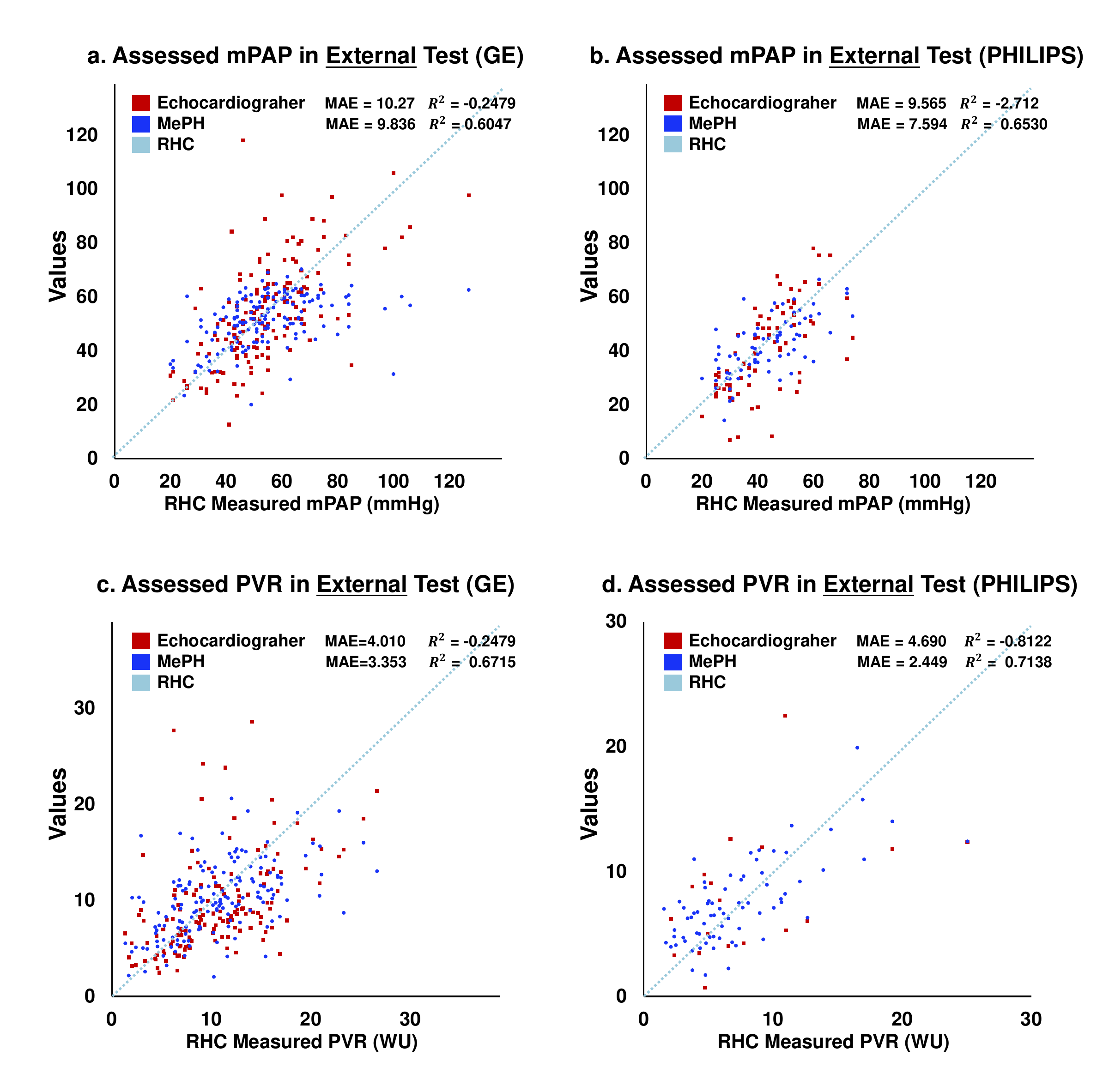}
\caption{\textbf{Performance comparison across different devices GE and PHILIPS}. \textbf{a,} The performance of mPAP assessment in the GE device. \textbf{b,} The performance of mPAP assessment in the PHILIPS device. \textbf{c,} The performance of PVR assessment in the GE device. \textbf{d,} The performance of PVR assessment in the PHILIPS device. Note that all results were reported by mean absolute error (MAE) and the coefficient of determination $R^2$ for PH classification based on mPAP (mmHg) and PH progression assessment based on PVR (WU). All results were reported in the external dataset.}
\label{fig:cross_device_result}
\end{suppfigure}
\clearpage

\newpage
\begin{suppfigure}[t]
\centering
\includegraphics[width=\textwidth]{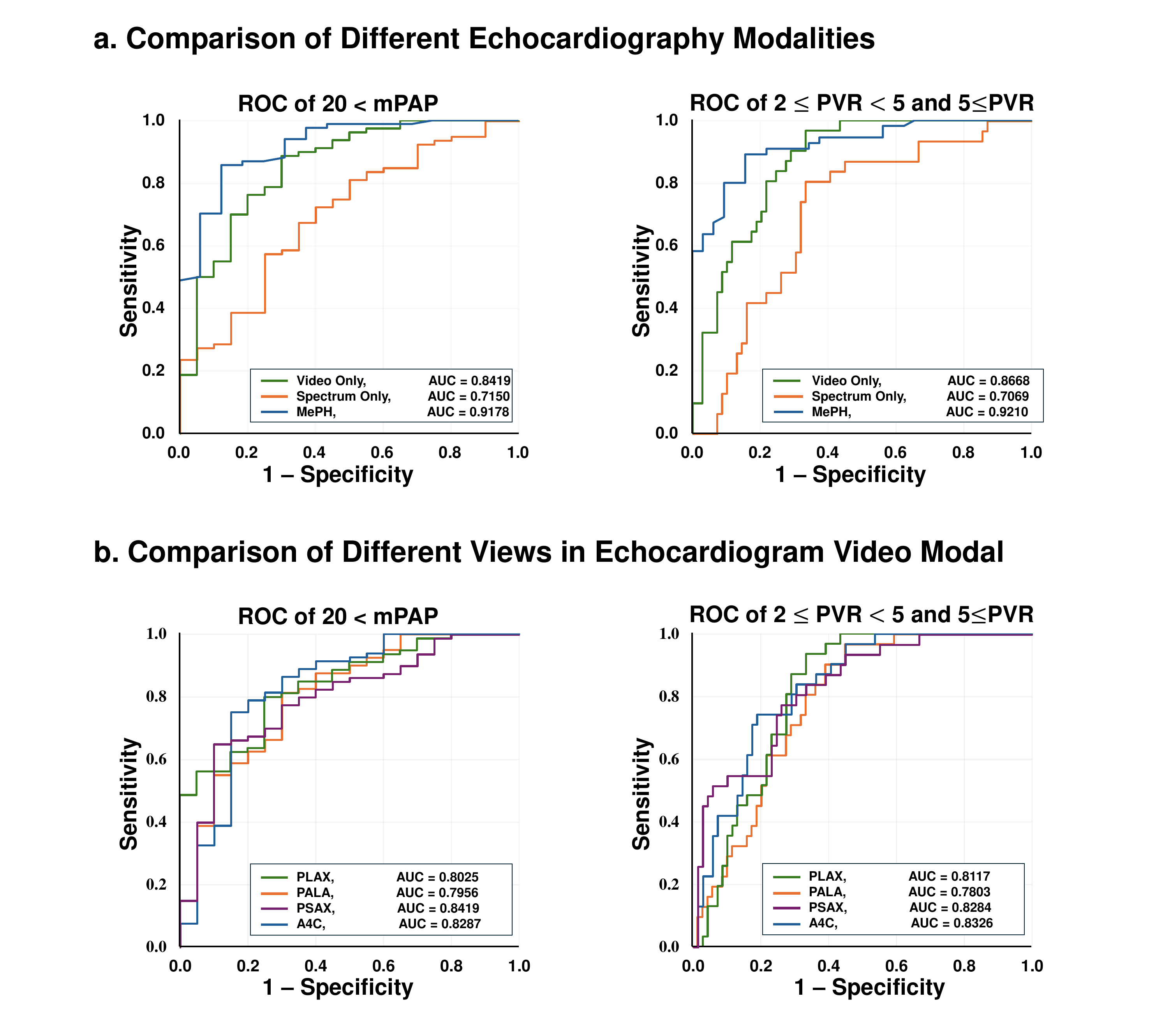}
\caption{\textbf{a, Performance comparison with different modalities: Echocardiogram videos, Doppler images and MePH}, results were reported by the area under the ROC curve as metric for PH classification based on mPAP and PH progression assessment based on PVR. \textbf{b, Performance comparison of different views in the echocardiogram video modal}, results were reported in PLAX, PALA, PSAX and A4C views. We use the area under the ROC curve as metric for PH classification based on mPAP and PH progression assessment based on PVR.}
\label{fig:ablation}
\end{suppfigure}
\clearpage

\newpage
\begin{suppfigure}[t]
\centering
\includegraphics[width=\textwidth]{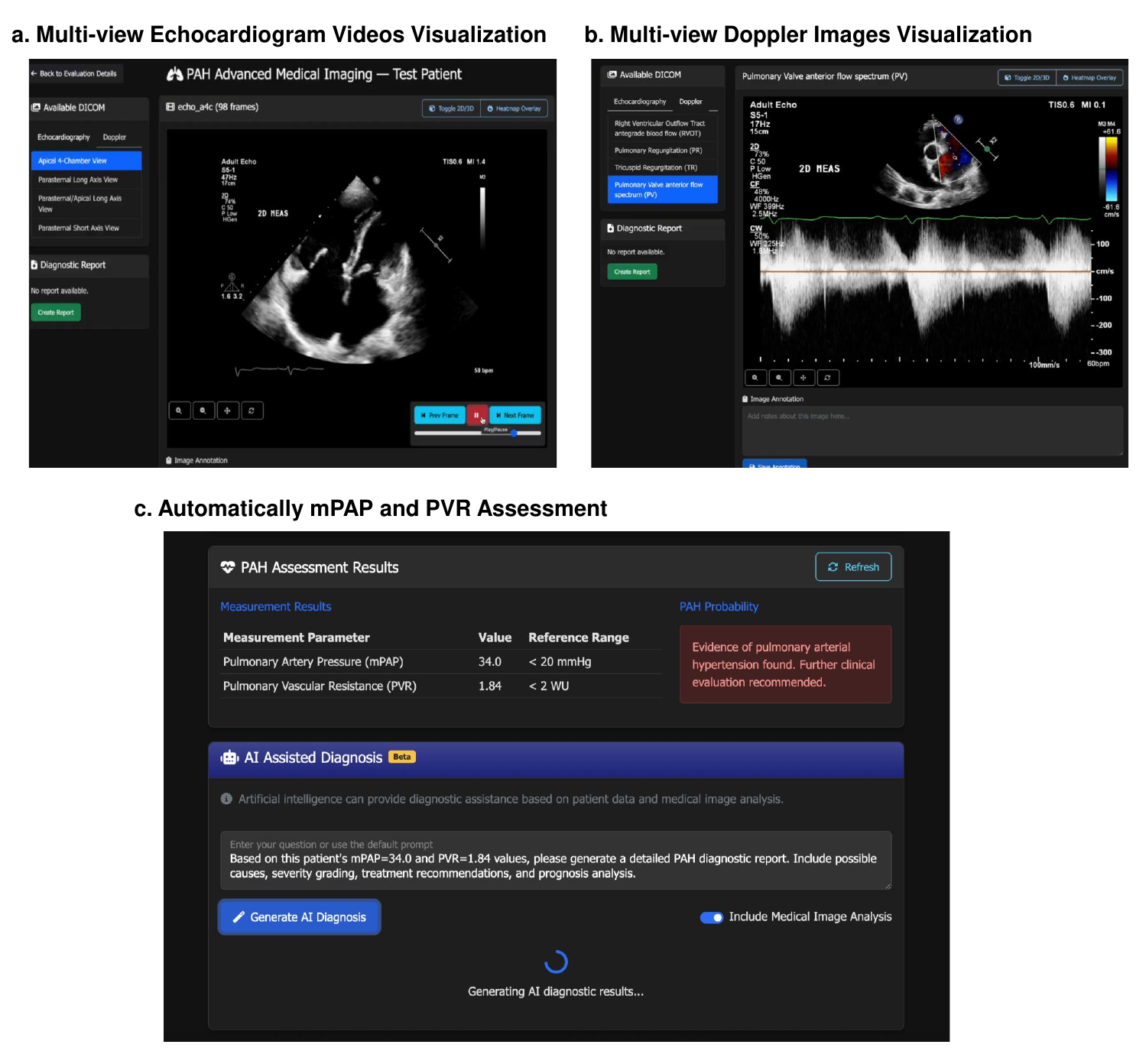}
\caption{\textbf{Visualization and AI-driven assessment workflow for pulmonary hypertension using multi-modal echocardiography.}
\textbf{a,} Multi-view echocardiogram video visualization demonstrates cardiac structures from different standard planes (e.g., Apical Four-Chamber view), enabling comprehensive assessment of right ventricular morphology and function.
\textbf{b,} Multi-view Doppler spectral image visualization captures hemodynamic flow patterns, such as the pulmonary valve anterior flow spectrum, to support non-invasive estimation of pulmonary pressure gradients.
\textbf{c,} Automated computation of mean Pulmonary Artery Pressure (mPAP) and Pulmonary Vascular Resistance (PVR) is shown, benchmarked against guideline-recommended reference ranges. The system integrates an AI-assisted diagnostic module that generates a probabilistic assessment of PAH and suggests clinical actions based on estimated hemodynamic parameters.}
\label{fig:demo}
\end{suppfigure}
\clearpage

\end{document}